# Chunking Strategies for Multimodal AI Systems

Shashanka B. R.[1], Mohith Charan[1], Seema Banu F[1]

{shashanka.b.r, mohith.charan, seema.banu}@ibm.com

## Abstract

Chunking has emerged as a critical technique that enhances generative models by grounding their responses in efficiently segmented knowledge [1]. While initially developed for unimodal (primarily textual) domains, recent advances in multimodal foundation models have extended chunking approaches to incorporate diverse data types, including images, audio, and video [2]. A critical component underpinning the success of these systems is the chunking strategy—how large, continuous streams of multimodal data are segmented into semantically meaningful units suitable for processing [3]. Despite its importance, chunking remains an under-explored area, especially in the context of multimodal systems where modality-specific constraints, semantic preservation, and alignment across modalities introduce unique challenges.

Our goal is to consolidate the landscape of multimodal chunking strategies, providing researchers and practitioners with a technical foundation and design space for developing more effective and efficient multimodal AI systems. This survey paves the way for innovations in robust chunking pipelines that scale with modality complexity, enhance processing accuracy, and improve generative coherence in real-world applications.

This survey provides a comprehensive taxonomy and technical analysis of chunking strategies tailored for each modality: text, images, audio, video, and cross-modal data. We examine classical and modern approaches such as fixed-size token windowing, recursive text splitting, object-centric visual chunking, silence-based audio segmentation, and scene detection in videos. Each approach is analyzed in terms of its underlying methodology, supporting tools (e.g., LangChain, Detectron2, PySceneDetect), benefits, and challenges, particularly those related to granularity-context trade-offs and multimodal alignment. Furthermore, we explore emerging cross-modal chunking strategies that aim to preserve alignment and semantic consistency across disparate data types [4]. We also include comparative insights, highlight open problems such as asynchronous information density and noisy alignment signals, and identify opportunities for future research in adaptive, learning-based, and task-specific chunking [5].

## 1. Introduction

The explosion of large-scale generative models such as GPT-4-turbo, PaLM, and Gemini has transformed the landscape of natural language understanding and generation. These foundation models (FMs) have demonstrated impressive capabilities across a wide range of tasks. However, even with expanding context windows (from 2K tokens in early models to 128K+ in recent versions), effective chunking remains essential for several reasons:

1. **Information Density Management**: While context windows have grown, not all information deserves equal attention. Chunking enables models to process content in semantically meaningful units instead of treating all content uniformly [8].
2. **Computational Efficiency:** Processing very long contexts can be computationally expensive; however, strategic chunking reduces redundant computations while preserving essential information [11].
3. **Attention Dilution Problem (Needle-in-a-Haystack):** As context length increases, attention tends to disperse across tokens, making it difficult to capture critical information. Structured chunking mitigates this by organizing content into semantically coherent segments that maintain focus [6].
4. **Structure Preservation:** Chunking utilizes the inherent organization present in raw data such as paragraphs, sections, or speaker turns to preserve structure and facilitate deeper comprehension [15].
5. **Memory Organization:** By reflecting how human cognition groups information into meaningful units, chunking enhances both processing efficiency and retrieval accuracy [9].

Early information processing systems were confined to text, but advances in multimodal foundation models now enable reasoning and generation across diverse modalities, including text, images, audio, and video. This progression opens new possibilities in areas such as medical diagnostics (text + image), surveillance (video + audio), education (lecture videos + slides), and enterprise analytics (documents integrated with charts, tables, and figures).

A central technical challenge in multimodal data lies in segmenting raw inputs into manageable, semantically coherent units that support effective processing. An effective chunking strategy ensures that information units preserve context, optimize relevance, and align with the modality-specific structures of the input [8]. Poor chunking can lead to fragmented information, semantic loss, processing noise, and ultimately, degraded generative quality. For example, in image chunking, indiscriminately dividing an image into patches may obscure object boundaries or discard spatial layout, while in audio, splitting at arbitrary intervals may break coherent speech units or disrupt speaker turns [3].

Despite its critical importance, the design of chunking strategies, especially in the multimodal context, remains under-explored. Most current systems adopt heuristic or ad hoc approaches that are modality-specific and often static. There exists a lack of systematic study, comparative analysis, and best practices around chunking techniques that generalize across modalities, tasks, and domains.

This survey aims to bridge that gap by providing a structured and detailed examination of chunking strategies tailored for multimodal AI systems. We organize our study around five core modalities:

1. **Text**: Token-based, sentence-level, paragraph-level, and semantic chunking techniques.
2. **Image**: Patch-based, object-centric, dense caption-based, and layout-aware segmentation.
3. **Audio**: Time-windowing, silence-based segmentation, and semantic speaker/topic-based chunking.
4. **Video**: Scene detection, keyframe summarization, and transcript-driven segmentation.
5. **Cross-modal**: Alignment-aware chunking across different modalities using embedding techniques and alignment models.

For each modality, we provide in-depth technical explanations, summarize tools and libraries used in practice (e.g., LangChain, Detectron2, Whisper, PySceneDetect), highlight open challenges, and cite state-of-the-art literature. We also propose a unified taxonomy for classifying chunking strategies and conclude by outlining future directions, including the promise of adaptive, neural, and learning-based chunking pipelines.

By consolidating diverse chunking strategies under a unified lens, this survey offers researchers and practitioners a comprehensive resource for designing, evaluating, and improving chunking components within multimodal AI architectures.

# 2. Background and Motivation

Chunking represents a fundamental information processing technique that segments large bodies of content into manageable, semantically coherent units. Originally conceptualized in cognitive psychology as a memory organization strategy, chunking has evolved into a critical component of modern Artificial Intelligence (AI) systems. Initially prominent in Natural Language Processing (NLP), the concept has since extended to other modalities as AI systems become increasingly multimodal.

A fundamental prerequisite for efficient information processing/retrieval is **chunking**, a concept that appears simple yet is technically indispensable. Raw, unstructured corpora are not directly amenable to computational operations; they must first be segmented into chunks or documents, self-contained units of information that balance contextual completeness with computational tractability [10]. These data units act as the fundamental building blocks that streamline subsequent processes such as model training, embedding generation, and inference.

While chunking strategies are relatively well established in the textual domain—commonly realized through techniques like token windowing, sentence or paragraph segmentation, recursive text partitioning [11], or semantic grouping based on topical coherence [20]—the rise of multimodal Artificial Intelligence (AI) systems introduces new layers of complexity:

1. **In Image-based processing**, the chunking task involves dividing complex visual input into semantically meaningful units that preserve object boundaries and layout information, often using tools like Visual Transformers, object detectors, or dense captioning models [12].
2. **In Audio processing**, segmentation must account for speaker turns, silence boundaries, and acoustic features while maintaining semantic continuity. Systems like Whisper and pyannote-audio support such segmentations.
3. **In Video processing**, temporal information and visual scene transitions must be coherently segmented, requiring multimodal cue fusion (video frames + audio + transcript) and the detection of scene or shot boundaries [13].
4. **In Cross-modal systems**, where a single document contains intertwined modalities (e.g., a news article with embedded images and audio clips), chunking must consider both intra-modality segmentation and inter-modality alignment [4].

Beyond traditional chunking techniques, Document Structure-Based Chunking has gained prominence by leveraging the inherent organization of documents (headings, sections, paragraphs) to define natural chunk boundaries [21]. This strategy maintains the logical flow of information but assumes a consistent document structure. Similarly, Sliding Window Chunking employs an overlapping approach where each chunk retains a portion of the previous one, preventing information gaps at chunk boundaries and ensuring sufficient context continuity [22].

The motivation for studying chunking strategies in depth is rooted in several practical and theoretical concerns:

1. **Processing Accuracy**: Poor chunking leads to noisy or incomplete results, reducing the quality of system outputs.
2. **Semantic Drift**: Arbitrary or fixed-length segmentation can break semantic units such as image regions, paragraphs, or conversations.
3. **Alignment Failures**: In cross-modal contexts, modality-specific chunking without synchronization can lead to mismatches (e.g., image chunks without corresponding text) [14].
4. **System Efficiency**: Chunking affects indexing, processing latency, and downstream compute requirements. Optimal chunk sizing helps balance granularity with system throughput.
5. **Task Relevance**: The optimal chunking strategy may vary by task (e.g., classification vs. generation) and by domain (e.g., legal vs. medical).

Despite its impact, chunking often remains an afterthought in system design, with many implementations relying on naive heuristics (e.g., fixed 512-token windows or arbitrary time-based splits) [5]. These methods fail to capture the nuanced structure and semantics of the data, particularly in multimodal settings. Moreover, there is little consensus or benchmarking across strategies, tools, or evaluation metrics.

This paper aims to fill that gap by systematically analyzing the landscape of chunking methods, particularly as they apply to the emerging class of multimodal AI systems. We categorize techniques, discuss their theoretical motivations, present practical implementation details, and outline key challenges and future directions—ultimately highlighting chunking as a first-class design concern in the next generation of AI architectures.

# 3. Taxonomy of Chunking Strategies

In the context of AI systems, especially multimodal ones, the chunking strategy defines how raw input data (text, image, audio, video, or a combination thereof) is partitioned into processable units. These chunks serve as the foundational elements for downstream tasks, typically represented via vector embeddings for efficient processing. A poorly designed chunking strategy can impair relevance, lead to information loss, or degrade multimodal alignment [3]. Therefore, a coherent taxonomy is essential to guide the selection, design, and evaluation of chunking methods.

We classify chunking strategies along five orthogonal axes:

## 3.1 By Modality

1. **Unimodal Chunking**: Chunking strategies applied independently to a single data type, such as:
    - **Text** (e.g., sentence or paragraph chunking),
    - **Image** (e.g., patch extraction),
    - **Audio** (e.g., silence detection).
2. **Cross-modal Chunking**: Chunking methods that consider inter-modality relationships, aligning, for example, image regions with captions or video scenes with subtitles [4].

## 3.2 By Granularity

1. **Fixed-size Chunking**: Segments are generated with a fixed number of tokens, frames, pixels, or seconds, regardless of semantic boundaries (e.g., 512-token blocks in text) [16]. This includes:
    - Non-Overlapping Fixed Chunks
    - Overlapping Fixed-Size Chunks
    - Sliding Window (Stride-Based)
    - Padding and Clipping Strategies

2. **Semantically Coherent Chunking**: Boundaries are determined by meaning-preserving units [17]. This includes:
    - Sentence-Level Segmentation
    - Paragraph-Level Segmentation
    - Topic Modeling-Based Segmentation
    - Embedding Similarity Clustering
    - TextTiling and Lexical Cohesion Methods
3. **Adaptive Chunking**: The chunk size varies depending on information density or semantic segmentation [5]. This includes:
    - Dynamic Chunking using Query-Driven Relevance
    - LLM-Informed Segmentation
    - Recursive Chunking
    - Mixture-of-Granularity (MoG)
4. **Document Structure-Based Chunking**: Leverages the inherent organization of documents such as headings, sections, and paragraphs to define chunk boundaries. This approach maintains the logical flow of information but assumes a clear and consistent document structure [21].

## 3.3 By Segmentation Heuristic

1. **Rule-based Segmentation**: Uses deterministic heuristics like sentence delimiters, silent pauses, or visual boundaries.
2. **Statistical/ML-based Segmentation**: Leverages topic modeling, clustering, or pretrained classifiers to determine chunk boundaries [18].
3. **Neural Segmentation:** Utilizes deep learning models to automatically identify chunk boundaries within data—for instance, by leveraging BERT-based embeddings, object detection frameworks such as YOLO or Detectron2, or scene segmentation models [19].
4. **Agentic Chunking:** Leverages Large Language Models (LLMs) to autonomously determine optimal segmentation points within a document based on semantic meaning and structural cues, effectively mimicking human reasoning when processing lengthy or complex texts [23].

## 3.4 By Alignment Awareness

1. **Modality-Agnostic Chunking:** Processes each modality independently, without enforcing synchronization across data types.
2. **Alignment-Aware Chunking:** Maintains semantic or temporal alignment between modalities [14]. Examples include:
    - **Text–Image Alignment:** Leveraging layout or HTML structure to align textual and visual content.
    - **Audio–Transcript Synchronization:** Ensuring that audio segments correspond accurately with their transcribed text.

3. **Multimodal Embedding Chunking:** Segments data based on shared multimodal embeddings learned from models such as CLIP, Flamingo, or GIT, thereby treating cross-modal signals as unified representations [7]. This includes:
   - **Joint Multimodal Embedding-Based Chunking** for integrated cross-modal understanding.
   - **Sliding Window Chunking:** Applies an overlapping segmentation strategy, where each chunk retains a portion of the preceding one. This approach minimizes information loss at boundaries and preserves contextual continuity [22].

## 3.5 By Purpose and Use Case

1. **Indexing Efficiency**: Chunking tuned for fast and sparse indexing (e.g., compressive chunking, low-resolution patching).
2. **Generative Utility**: Chunking optimized for providing maximal helpful context to generators (e.g., overlap-aware recursive chunking).
3. **Alignment for Processing Recall**: Strategies that maximize hit rate and semantic precision (e.g., text-anchored vision chunking for OCR documents).
4. **Metadata-Augmented Chunking**: Enhances processing by tagging chunks with metadata such as headings, timestamps, and document types, providing additional context to the AI system [24].
5. **Hybrid Chunking**: Combines elements of different chunking strategies to leverage the advantages of each, allowing for tailored approaches based on the specific needs of the application [25].

**Table 1**: Taxonomy of Chunking Strategies

| **Axis** | **Type** | **Description / Example** |
|---|---|---|
| **1. By Modality** | Unimodal Chunking | Chunking per data type (e.g., text, image, audio) |
| | Text | Sentence/paragraph splitting |
| | Image | Patch extraction |
| | Audio | Silence-based segmentation |
| | Cross-modal Chunking | Aligns across modalities (e.g., video + subtitles, image + caption) |
| **2. By Granularity** | Fixed-size Chunking | Fixed-size segments (e.g., 512-token blocks) |
| | Semantic Chunking | Meaning-based units (e.g., topics, scenes, paragraphs) using AI/embeddings |
| | Adaptive Chunking | Variable size based on information density or content complexity |
| | Document Structure-Based | Uses headings/sections/paragraphs to define chunks |

| **3. By Segmentation Heuristic** | Rule-based Segmentation | Uses fixed rules (e.g., sentence boundaries, silent pauses) |
|---|---|---|
| | Statistical/ML-based | Uses models like topic modeling, clustering, or classifiers |
| | Neural Segmentation | Deep learning-based (e.g., BERT, YOLO, scene detection) |
| | Agentic Chunking | LLM-guided semantic segmentation simulating human-like reasoning |
| **4. By Alignment Awareness** | Modality-agnostic Chunking | Modalities chunked independently with no alignment |
| | Alignment-aware Chunking | Maintains semantic/temporal sync across modalities (e.g., captions to images) |
| | Multimodal Embedding Chunking | Joint chunking based on multimodal embeddings (e.g., CLIP, Flamingo) |
| | Sliding Window Chunking | Overlapping chunks to maintain context and prevent gaps |
| **5. By Purpose / Use Case** | Indexing Efficiency | Lightweight chunks optimized for fast retrieval (e.g., compressive or sparse representations) |
| | Generative Utility | Chunks designed to feed helpful context to LLMs |
| | Alignment for Processing Recall | Enhances accuracy, e.g., with modality alignment |
| | Metadata-Augmented Chunking | Adds tags like timestamps, headings, or type for better processing |
| | Hybrid Chunking | Blends multiple strategies for flexibility and domain-specific optimization |

This taxonomy provides a comprehensive lens to evaluate, compare, and select chunking strategies tailored to specific modalities, system architectures, and downstream tasks. In the following sections, we apply this taxonomy to examine chunking strategies across Text, Image, Audio, Video, and Cross-modal AI systems.

# 4. Text Chunking

Text chunking segments textual data into discrete units that optimize semantic coherence, context preservation, and computational efficiency. These units are foundational for NLP tasks, including indexing, embedding, retrieval, and generation, particularly in Retrieval-Augmented Generation (RAG), semantic search, question-answering, and summarization. Effective chunking ensures segments are meaningful, retrievable, and processable, balancing granularity and context. Text chunking strategies are categorized into three paradigms: fixed-size chunking, semantically coherent chunking, and adaptive chunking. Below, we provide a detailed analysis of each paradigm and its sub-methods, with mathematical formulations, practical considerations, and trade-offs.

### 4.1.1 Fixed-Size Chunking

Fixed-size chunking partitions text into uniform segments, typically defined in terms of tokens, characters, or words. This approach prioritizes simplicity, predictability, and computational efficiency, making it particularly suitable for large-scale applications where throughput and scalability are paramount. However, its disregard for semantic boundaries can lead to disruptions in contextual coherence, which in turn may reduce retrieval accuracy in tasks requiring fine-grained understanding [16]. Within this framework, four primary variants are typically employed: non-overlapping fixed chunks, overlapping fixed-size chunks, stride-based sliding windows, and padding or clipping strategies.

#### 4.1.1.1 Non-Overlapping Fixed Chunks

This method divides a text sequence into consecutive, non-redundant segments of fixed length, ensuring that each token belongs to exactly one chunk. It is highly computationally efficient, with minimal storage overhead due to the absence of duplication, making it well suited for indexing large corpora or preprocessing data for embedding models constrained by fixed input lengths.

However, this approach disregards linguistic boundaries such as sentences or paragraphs, which can fragment coherent ideas and reduce recall for queries spanning multiple chunks [11]. For example, if a concept is split across two adjacent chunks, retrieval systems may fail to capture it entirely. The chunk size ($K$) thus becomes a critical hyperparameter: smaller values produce finer-grained but context-poor segments, whereas larger values yield unwieldy chunks that can hinder retrieval or downstream processing. In practice, short texts (shorter than $K$) collapse into a single chunk, while longer texts often end with a residual segment containing fewer tokens.

**Practical Considerations:** Non-overlapping chunking is implemented in widely used tools such as *LangChain's CharacterTextSplitter* and *Hugging Face tokenizers*, which allow splitting by token or character counts [5]. The choice of chunk size ($K$) is typically task-dependent: smaller chunks (e.g., 100 tokens) are advantageous for fine-grained retrieval, while larger chunks (e.g., 512 tokens) align better with transformer model input constraints. This method works best for structured or well-delineated text, but it performs less effectively for narratives or dialogues, where semantic dependencies frequently extend across chunk boundaries.

**Trade-offs:** This approach offers strong computational efficiency, with storage requirements proportional to $N$ and linear processing time, making it ideal for large-scale datasets. It preserves precision for queries fully contained within a single chunk, but sacrifices recall when context spans multiple segments. Consequently, it is most suitable for applications where computational speed and scalability are prioritized over semantic continuity, such as large batch embedding generation.

#### 4.1.1.2 Overlapping Fixed-Size Chunks

This method addresses the boundary-fragmentation issue inherent in non-overlapping chunking by introducing controlled redundancy, wherein consecutive chunks share a subset of tokens. Such overlap ensures that information appearing near chunk boundaries is replicated across adjacent segments, thereby improving the likelihood of retrieving relevant context for queries that span multiple chunks. Although this strategy enhances recall and preserves contextual continuity, it does so at the expense of increased storage requirements and higher computational overhead resulting from content duplication. This characteristic is particularly beneficial in Retrieval-Augmented Generation (RAG) pipelines, where maintaining contextual continuity directly contributes to more coherent and contextually relevant responses [27]. The overlap size (O) $\left(where(0 \leq O < K)\right)$ determines redundancy: larger overlaps retain more contextual information but incur higher storage and processing costs, whereas smaller overlaps behave similarly to non-overlapping chunking with minimal redundancy. Overlapping chunks can lead to redundant retrieval results, potentially lowering precision if irrelevant segments are retrieved. The method is sensitive to the choice of (K) and (O), which must balance context preservation against computational overhead.

**Practical Considerations**: Supported by tools like LangChain's `CharacterTextSplitter` with an overlap parameter or Hugging Face's tokenizers with custom strides, this method is widely used in document retrieval and question answering. Choosing (O) (e.g., 10–25% of (K) requires experimentation to optimize recall without excessive redundancy. Edge cases include texts shorter than (K), producing a single chunk, or large (O), which significantly increases chunk count.

**Trade-offs**: Overlapping chunking improves recall by preserving cross-boundary context, making it suitable for tasks requiring continuity, like RAG. However, storage increases by a factor of, $\left(K/(K - O)\right)$, and redundancy may reduce precision. It is less efficient than non-overlapping chunking but more robust for semantic tasks [27].

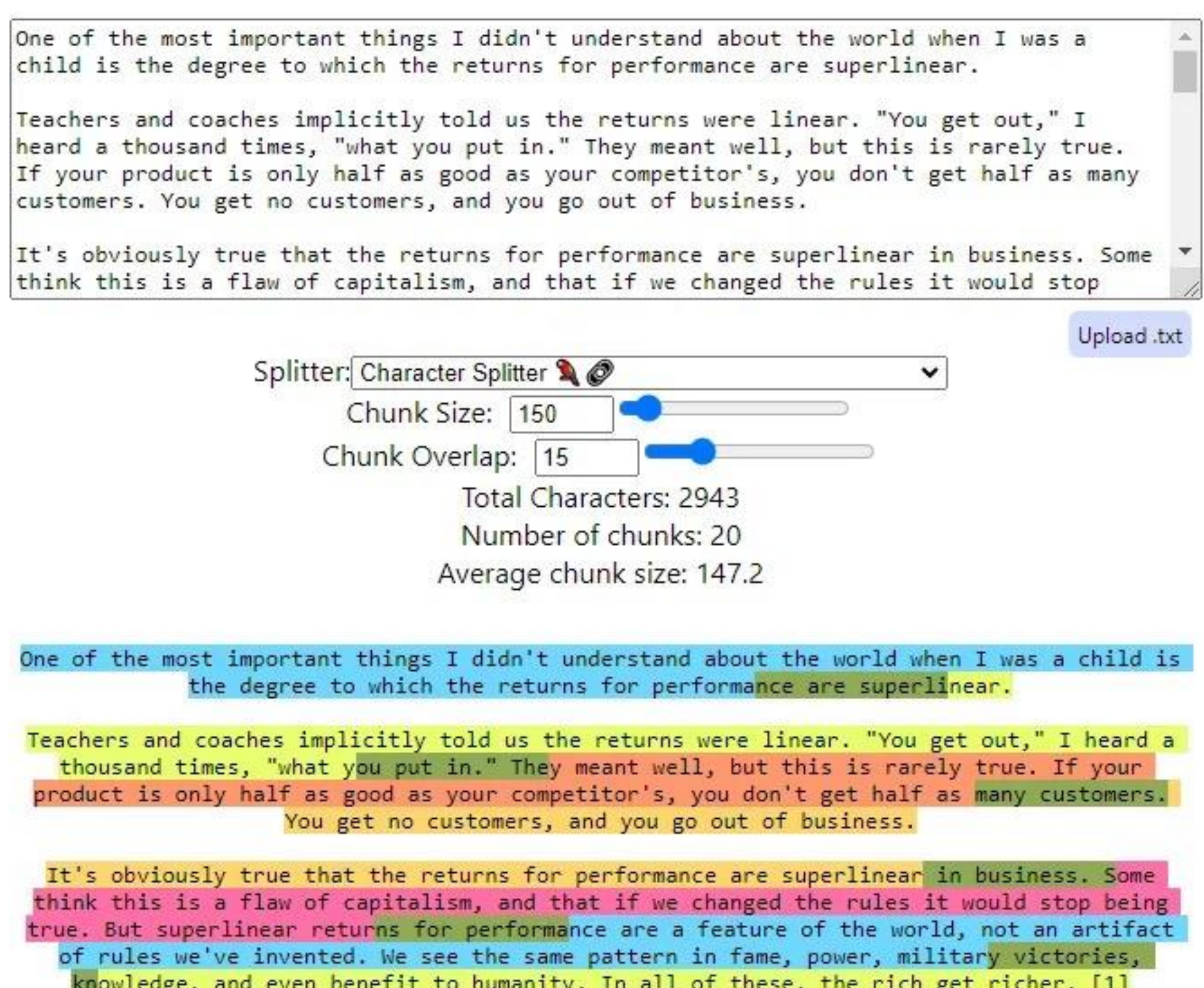

Figure 1: Overlapping Fixed-Size Chunks [15]

#### 4.1.1.3 Sliding Window (Stride-Based) Chunking

Sliding window chunking generalizes overlapping chunking by defining a window size (K) and stride $\left(S\ (S \leq K)\right)$, where the stride determines the step size between chunks. The overlap is (K - S), allowing fine-grained control over redundancy and coverage. When (S = K), it reduces to non-overlapping chunking; when (S < K), it introduces overlap; (S > K) (rare) skips tokens. This method is prevalent in transformer-based models for processing long sequences, maintaining contextual dependencies across token spans [22]. It is effective for sequential tasks like summarization or long-context question answering, where local context is crucial. The stride (S) balances redundancy against computational cost: smaller strides increase context but generate more chunks, while larger strides reduce redundancy but risk context loss. Edge cases include texts shorter than (K), yielding one chunk, or very small (S), producing excessive overlap.

**Mathematical Formulation**: For window size (K) and stride (S), the $\left((i) - th\right)$ chunk is:

[

$$C_i = \left[t_{(i-1)S+1}, t_{(i-1)S+2}, \dots, t_{\min((i-1)S+K,N)}\right], \quad i = 1,2,\dots,\lceil (N-K)/S \rceil + 1$$

]

The start index is $((i-1)S+1)$, and the chunk contains (K) tokens unless truncated. The number of chunks is$(\lceil (N-K)/S \rceil + 1)$, and the overlap is (K - S). For (K = 200), (S = 150), and (N = 1000), chunks are 1–200, 151–350, 301–500, etc., with 50-token overlaps, yielding 6 chunks.

**Practical Considerations**: Frameworks such as *PyTorch*, *TensorFlow*, and *Hugging Face tokenizers* provide native support for sliding-window chunking with configurable stride parameters. This method is well-suited for sequential processing tasks but requires careful tuning of the chunk size (K) and stride (S). A common configuration is $S$= 0.75K, which balances redundancy with efficiency by introducing moderate overlap. However, sliding-window approaches are less effective for texts with abrupt topical shifts, where semantic or adaptive chunking methods provide superior boundary alignment.

**Trade-offs**: Sliding window chunking offers flexibility in balancing context and redundancy. Smaller strides enhance recall but increase computational and storage costs, while larger strides reduce costs but risk context loss. It is efficient for sequential tasks but requires careful parameter tuning [22].

#### 4.1.1.4 Padding and Clipping Strategies

Padding and clipping handle cases where text does not divide evenly into fixed-size chunks or when uniform input sizes are required for model processing. Padding involves appending neutral tokens (e.g., <PAD>) to reach a fixed chunk size, ensuring compatibility with transformer-based models that require uniform input lengths. This technique is widely used in batched processing but can introduce minor noise, potentially affecting embedding quality or retrieval relevance. Clipping, on the other hand, removes excess tokens that exceed the defined chunk size or truncates the final segment, thereby reducing redundancy but at the risk of information loss. The choice between padding and clipping is largely task-dependent: padding is typically favored during model training or inference, whereas clipping is more suitable for storage-constrained indexing scenarios. Edge cases include texts shorter than *K*, which require padding to meet length requirements, and longer texts with residual segments, where clipping decisions must balance completeness against efficiency.

**Mathematical Formulation**:

1. **Padding**: For a chunk $(C_i)$ with (M < K) tokens:

[

$$C_i = [t_1, t_2, \ldots, t_M, , \ldots, ], \quad \text{with } K - M \text{ padding tokens}$$

]

2. **Clipping**: For a chunk with (M > K) tokens, retain only the first (K):

[

$$C_i = [t_1, t_2, \ldots, t_K]$$

]

Alternatively, for the final chunk with $(M \ \leq K)$:

[

$$C_i = [t_{N-M+1}, t_{N-M+2}, \ldots, t_N]$$

]

**Practical Considerations:** Padding and clipping are widely supported across modern deep learning frameworks and text processing libraries, including *LangChain* [29]. Padding is essential for generating uniform-length inputs in batched transformer models, while clipping is typically employed to maintain storage efficiency during indexing. The choice of padding token (e.g., <PAD> or zero vectors) is model-dependent and generally determined by the tokenizer in use. Clipping strategies can also be configured to retain either the beginning or the end of a segment, depending on which portion of the content carries greater relevance.

**Trade-offs:** Padding ensures model compatibility but may introduce non-informative tokens, potentially reducing precision. Conversely, clipping minimizes redundancy but at the cost of discarding valuable information, which can lower recall. Both techniques are computationally efficient; however, their optimal use requires task-specific tuning to balance the trade-off between data integrity and processing efficiency.

### 4.1.2 Semantically Coherent Chunking

This approach aims to align chunk boundaries with meaningful linguistic or topical units—such as sentences, paragraphs, or coherent themes—thereby preserving contextual integrity and improving retrieval performance. These methods leverage syntactic structures, statistical indicators, or semantic representations to maintain coherence across segments, which is particularly crucial for tasks requiring deep semantic understanding [17]. Common strategies include sentence- or paragraph-level segmentation, topic-model-based partitioning, clustering driven by embedding similarity, and algorithms such as *TextTiling*, which utilize lexical cohesion to identify natural discourse boundaries.

#### 4.1.2.1 Sentence-Level Segmentation

This method treats individual sentences as discrete chunks, identifying boundaries through punctuation or syntactic cues. By regarding sentences as the minimal units of discourse, it achieves high precision, as each chunk typically represents a single, self-contained idea. Sentence-level segmentation is particularly effective for tasks such as question answering, where queries often target specific statements. However, its limited contextual scope can reduce recall for queries that span multiple sentences. Variable sentence lengths also complicate model inputs, often requiring padding or truncation for uniformity. The method is sensitive to text quality: poorly punctuated or informal content—such as social media posts—can yield unreliable boundaries. Edge cases include single-sentence texts, which form only one chunk, and run-on sentences, which may produce disproportionately large ones [11].

**Practical Considerations:** Natural Language Processing (NLP) libraries such as NLTK and spaCy provide sentence tokenization tools based on either rule-driven heuristics or machine learning models. spaCy's *sentencizer*, in particular, performs robustly across multiple languages. While sentence-level tokenization is computationally efficient, it often requires preprocessing to handle noisy or unstructured text. Its effectiveness tends to decline in narrative or dialogue-heavy documents, where meaningful context frequently extends beyond sentence boundaries.

**Trade-offs:** Sentence-level segmentation prioritizes precision by isolating self-contained ideas but sacrifices recall due to its limited contextual window. It remains computationally lightweight and interpretable yet struggles with variable sentence lengths and complex information that spans multiple sentences. Consequently, this approach is best suited for fine-grained retrieval or localized understanding tasks [11].

#### 4.1.2.2 Paragraph-Level Segmentation

Paragraph-level segmentation defines chunk boundaries based on formatting cues such as line breaks, indentation, or paragraph tags, grouping semantically related sentences into cohesive units. By capturing broader contextual information than sentence-level segmentation, this approach enhances retrieval recall and comprehension for queries requiring multi-sentence reasoning—such as document-level search, summarization, or question answering. However, paragraph lengths can vary significantly across documents, and many unstructured sources—such as web pages, OCR outputs, or social media text—may lack reliable paragraph markers. Moreover, the method implicitly assumes thematic coherence within paragraphs, which may not hold in poorly structured or automatically generated content [21]. Edge cases include single-paragraph documents, which yield only one chunk, and very short paragraphs, which can lead to overly fine-grained segmentation.

**Practical Considerations:** In practice, tools like *LangChain's RecursiveCharacterTextSplitter* commonly use double newline delimiters (\n\n) to detect paragraph boundaries, while libraries such as *BeautifulSoup* identify <p> tags in HTML documents. This approach performs well for well-formatted text but often requires fallback mechanisms when applied to noisy or inconsistently structured inputs. Preprocessing steps—such as standardizing line breaks, removing extraneous whitespace, and repairing incomplete markup—can significantly improve segmentation reliability and coherence.

**Trade-offs**: Paragraph-level segmentation improves recall by capturing context but is sensitive to formatting inconsistencies. It maintains moderate precision and is efficient, making it suitable for thematic retrieval tasks, though variable sizes pose challenges for model inputs [21].

#### 4.1.2.3 Topic Modeling-Based Segmentation

Topic modeling-based segmentation, often using Latent Dirichlet Allocation (LDA), segments text by identifying thematic shifts. The text is divided into windows (fixed or sliding), each assigned a topic distribution. Boundaries are placed where topic distributions diverge significantly, ensuring topical coherence within chunks. This is ideal for long documents with distinct sections, enhancing retrieval relevance for theme-specific queries [20]. However, it is computationally intensive, requiring topic model training and hyperparameter tuning (e.g., number of topics (K)). Performance depends on text quality and model accuracy, with noisy or short texts yielding unreliable segments. Edge cases include single-topic texts, producing one chunk, or overly granular models, creating fragmented segments.

**Practical Considerations**: Libraries like Gensim or scikit-learn support LDA-based segmentation, with preprocessing (e.g., stopword removal, lemmatization) critical for accuracy [20]. The choice of (K) and window size affects segmentation quality, requiring experimentation. The method is robust for academic or technical texts but less effective for short or informal data.

**Trade-offs**: Topic modeling enhances precision and recall by ensuring topical coherence but is computationally expensive and sensitive to model parameters. It is ideal for thematic retrieval but impractical for real-time applications or resource-constrained settings [20].

#### 4.1.2.4 Embedding Similarity Clustering

Embedding-based chunking uses dense vector representations of sentences or paragraphs (e.g., from BERT or Sentence Transformers) to cluster semantically similar units. Boundaries are placed where similarity between consecutive units drops below a threshold, or clustering algorithms group similar units into chunks. This dynamically adapts chunk sizes to content, balancing precision and recall by ensuring semantic coherence. It is effective for texts with variable topic granularity, such as technical reports or narratives, but requires pre-trained embedding models and significant computational resources. Edge cases include highly similar texts, producing large chunks, or dissimilar sentences, yielding fragmented segments.

**Mathematical Formulation**: For sentences $(S_1, S_2, \ldots, S_N)$ with embeddings $(e_1, e_2, \ldots, e_N)$, a new chunk starts when cosine similarity falls below a threshold (τ):

[

$$\cos(e_i, e_{i+1}) = \frac{e_i \cdot e_{i+1}}{|e_i||e_{i+1}|} < \tau$$

]

Chunks are:

[

$$C_j = \left[S_{b_{j-1}+1}, \ldots, S_{b_j}\right], \quad \text{where } b_j \text{ are indices with } \cos\left(e_{b_j}, e_{b_j+1}\right) < \tau\tau$$

]

Alternatively, clustering (e.g., hierarchical or DBSCAN) groups sentences into chunks based on embedding proximity.

**Practical Considerations**: Tools like LangChain's `SemanticTextSplitter` or Sentence Transformers (`all-MiniLM-L6-v2`) support embedding-based chunking [29]. The choice of embedding model and (τ) affects segmentation quality, with higher (τ) producing finer chunks. Preprocessing to normalize text enhances embedding reliability. The method is robust for diverse texts but computationally heavy.

**Trade-offs**: Embedding-based chunking maximizes semantic coherence, improving precision and recall, but is resource-intensive due to embedding computation. It is ideal for semantic retrieval but less suitable for large-scale or real-time applications.

#### 4.1.2.5 Late Chunking

Late Chunking reverses the conventional chunking pipeline by first encoding the entire document into token-level embeddings, and only then partitioning these embeddings into chunks. This stands in contrast to traditional strategies that segment the raw text prior to embedding. By deferring chunking until after the semantic representation is formed, each chunk retains global context, enabling stronger semantic coherence and long-range dependency modeling [39].

This approach leverages long-context embedding models, such as Jina-embeddings-v2, which can process up to 8,192 tokens in a single pass. These models output contextualized token embeddings that are already informed by the entire document. Once these embeddings are obtained, chunk boundaries are applied post hoc using sentence-level or paragraph-level heuristics. Each resulting chunk vector is computed via pooling (typically mean pooling) over the embeddings within that boundary [39].

The key advantage of Late Chunking lies in its context-aware chunk embeddings. Since each token is influenced by the surrounding document, this method preserves references, pronouns, and thematic consistency that are typically lost when texts are split early. For example, in a Wikipedia article about Berlin, a sentence like *"Its more than 3.85 million inhabitants…"* becomes ambiguous when separated from its subject, *"Berlin."* Late Chunking prevents such detachment by keeping references semantically grounded, ensuring that entities and their descriptions remain contextually linked. This alignment improves both retrieval accuracy and generative relevance [39].

Late Chunking is particularly effective in retrieval-augmented generation (RAG) systems that process long documents, where reasoning often extends across chunk boundaries. By maintaining semantic continuity, it allows embeddings to capture document-level meaning rather than isolated fragments. This makes the approach especially valuable in domains such as biomedical literature retrieval, legal document analysis, and policy-oriented question answering. It also enhances LLM-based summarization, where the coherence of input segments directly influences the quality and factual consistency of the generated summaries [39].

**Practical Considerations:** Late Chunking assumes access to transformer models with long-context capacity and the ability to expose intermediate token embeddings. No model retraining is required. However, the approach introduces computational and memory overhead, as it necessitates processing full documents prior to segmentation. Chunk boundaries still need to be defined (e.g., using regular expressions, sentence detectors, or markup tags), but they are applied in the embedding space, not the raw text space.

**Trade-offs:** Late Chunking yields superior semantic alignment and retrieval performance, as shown in benchmarks like SciFact (+1.9% nDCG@10), TRECCOVID (+1.34%), and NFCorpus (+6.52%). However, the method is less efficient for real-time processing or resource-constrained environments, and its effectiveness degrades when applied to models with short context limits. It also assumes that documents fit within the model's maximum token window, or that appropriate segmentation logic exists to embed them in overlapping blocks [39].

#### 4.1.2.6 TextTiling and Lexical Cohesion Methods

TextTiling segments text by analyzing lexical cohesion, measuring word overlap or similarity between adjacent blocks (pseudo-sentences or windows). Boundaries are placed at points of low similarity, indicating shifts in lexical content. This is lightweight and interpretable, relying on surface-level features, but less robust to synonyms or paraphrasing compared to embedding-based methods [28]. It performs well on structured texts like articles but struggles with informal or noisy data (e.g., social media). The method is sensitive to window size and similarity thresholds, with small windows producing fragmented chunks and large windows missing subtle shifts. Edge cases include highly cohesive texts, yielding few chunks, or disjointed texts, producing excessive segments.

**Mathematical Formulation**: For pseudo-sentences $(P_1, P_2, \ldots, P_M)$, with word frequency vectors $(v_i)$, similarity is computed as:

[

$$\text{sim}(P_i, P_{i+1}) = \frac{v_i \cdot v_{i+1}}{|v_i||v_{i+1}|}$$

]

Chunks are:

[

$$C_j = \left[P_{b_{j-1}+1}, \ldots, P_{b_j}\right], \quad \text{where } b_j \text{ are indices with sim}\left(P_{b_j}, P_{b_j+1}\right) < \tau\tau$$

]

The number of chunks depends on detected boundaries, influenced by the threshold $(\tau)$.

**Practical Considerations**: NLTK's TextTiling implementation supports lexical cohesion-based segmentation, with configurable window sizes and thresholds [28]. Preprocessing (e.g., tokenization, stopword removal) enhances accuracy. The method is efficient for structured texts but requires tuning for noisy data.

**Trade-offs**: TextTiling is computationally efficient and easy to interpret, offering a balanced trade-off between precision and recall. However, it tends to struggle with capturing subtle semantic transitions, making it better suited for the rapid segmentation of well-structured texts rather than complex or informal content [28].

### 4.1.3 Adaptive Chunking

Adaptive chunking dynamically adjusts segmentation boundaries based on content characteristics, query intent, or task-specific objectives, using AI-driven strategies to optimize retrieval performance [5]. Common approaches include query-driven relevance chunking, LLM-informed segmentation, recursive chunking, and Mixture-of-Granularity (MoG)methods.

#### 4.1.3.1 Dynamic Chunking using Query-Driven Relevance

Query-driven chunking performs segmentation dynamically at query time, identifying and extracting text spans most relevant to a specific query. Relevance scores—computed at the sentence or span level using models such as BM25 or neural retrievers—guide the selection of high-scoring regions, which are often expanded to include adjacent contextual content. This adaptive behavior maximizes precision by tailoring chunks directly to the information need, making it particularly valuable for question-answering and targeted retrieval applications.

However, this approach incurs additional query-time latency since segmentation occurs on-the-fly, and it may overlook broader contextual information, reducing recall for complex or multi-faceted queries. Its performance strongly depends on the quality of the underlying relevance scoring and indexing mechanisms that enable efficient dynamic extraction. Edge cases include queries that yield no relevant spans, producing empty chunks, or overly broad queries that result in excessively large segments.

**Mathematical Formulation**: For a query (Q) and document (D) with sentences $(S_1, S_2, \dots, S_N)$,a relevance score $\left(R(Q, S_i)\right)$is computed. Chunks are:

[

$$C_j = \left[S_{b_{j-1}+1}, \dots, S_{b_j}\right], \quad \text{where } b_j \text{ are indices with } R\left(Q, S_{b_j}\right) > \tau$$

]

The number of chunks depends on the query and threshold (\tau), with expansion rules (e.g., include $(\pm k)$ sentences) defining chunk size.

**Practical Considerations:** Frameworks such as Haystack and Elasticsearch support query-driven chunking by integrating coarse-grained retrieval methods (e.g., BM25) with fine-grained span extraction. Neural retrievers like Dense Passage Retriever (DPR) can further enhance relevance scoring by capturing deeper semantic relationships between queries and documents. While this approach requires pre-indexing sentences or spans introducing additional setup complexity, it enables dynamic, query-specific adaptation, significantly improving precision in targeted retrieval scenarios.

**Trade-offs:** Query-driven chunking prioritizes precision for specific queries but introduces latency due to on-demand segmentation, which may constrain its use in real-time or high-throughput systems. Although it enhances focused retrieval, it can reduce recall for queries requiring broader contextual reasoning. Effective deployment, therefore, requires careful calibration of relevance thresholds to maintain an optimal balance between precision and contextual coverage.

#### 4.1.3.2 LLM-Informed Segmentation

LLM-driven chunking capitalizes on the contextual awareness and semantic reasoning capabilities of large language models to determine optimal chunk boundaries. Instead of relying on static heuristics or fixed token windows, this approach prompts an LLM to identify natural points of topical or thematic transition within the text. For instance, the model can be instructed to insert separators where the narrative shifts focus, ensuring that each resulting segment reflects a coherent and self-contained idea. This method effectively mirrors human judgment in text segmentation, enabling contextually adaptive and semantically rich chunk formation that enhances retrieval and generation quality. This approach is particularly effective for complex texts with subtle shifts, improving both precision and recall [23]. However, it is computationally intensive, requiring substantial processing resources or API access, and is sensitive to prompt design; ambiguous instructions can result in inconsistent chunk boundaries. Edge cases include very short texts, which yield few chunks, and overly detailed prompts, which may generate excessively fragmented segments.

**Practical Considerations**: LangChain supports LLM-based splitters, using models like OpenAI's or local deployments. Prompt engineering is critical, with clear instructions (e.g., "Segment into topical sections") improving consistency. The method is robust for high-quality segmentation but impractical for large-scale or low-resource settings.

**Trade-offs**: LLM-informed segmentation produces high-quality, context-aware chunks, maximizing precision and recall, but is resource-intensive and prompt-sensitive. It is ideal for complex texts but less suitable for real-time or resource-constrained applications [23].

#### 4.1.3.3 Recursive Chunking

Recursive chunking applies a hierarchy of splitting rules, starting with large units (e.g., paragraphs) and subdividing if they exceed a size limit (K). It prioritizes semantic units, falling back to finer-grained splits (e.g., sentences, characters) as needed, balancing coherence and manageability [5]. It is computationally efficient, as it avoids complex modeling, but requires defining a splitter sequence and size threshold. Edge cases include texts smaller than (K), yielding one chunk, or highly nested structures, increasing recursion depth.

**Mathematical Formulation**: For a text (T) and splitters $(f_1, f_2, \ldots, f_n)$, chunks are generated recursively:

[

$$C_i = f_j(T_i), \quad \text{if } |T_i| > K, \quad j = 1,2,\ldots,n$$

]

Each splitter $(f_j)$ produces segments, and the process repeats until all chunks satisfy $(|C_i| \leq K)$. The number of chunks depends on the text and splitters.

**Practical Considerations**: LangChain's `RecursiveCharacterTextSplitter` supports this, prioritizing delimiters like `\n\n`, `\n`, or punctuation [29]. The method is widely used in RAG pipelines for its flexibility. The choice of (K) and splitter order affects segmentation quality, with experimentation needed for optimal results.

**Trade-offs**: Recursive chunking is efficient and robust, balancing precision and recall across text types, but requires careful splitter design. It is suitable for general-purpose chunking but may not capture subtle semantic shifts as effectively as model-based methods [5].

#### 4.1.3.4 Mixture-of-Granularity (MoG)

Mixture-of-Granularity (MoG) maintains chunks at multiple granularities (e.g., sentence, paragraph, section) and selects the optimal level at query time using a router model or heuristic. This adapts chunk size to the query's scope, enhancing precision for specific queries and recall for broad ones [25]. It is ideal for dynamic retrieval scenarios with variable query types, but increases indexing complexity, as multiple chunk versions must be stored. The router's accuracy is critical, with poor predictions degrading performance. Edge cases include queries with ambiguous granularity, leading to suboptimal chunk selection, or short texts, reducing granularity options.

**Mathematical Formulation**: For a query (Q) and document (D), the router predicts the optimal granularity (G):

[

$$G = \arg\max_{g} P\,(g|Q, D)$$

]

Chunks are selected from the set $(C_g)$, where $(C_g)$ are chunks at granularity (g) (e.g., sentence, paragraph). The number of chunks depends on the chosen granularity.

**Practical Considerations**: MoG is an emerging approach, supported by custom RAG pipelines or LangChain with multi-granularity indexing [29]. The router may use query length, type, or relevance scores to select granularity. The method requires robust indexing infrastructure to manage multiple chunk sets.

**Trade-offs**: MoG offers flexibility, optimizing precision and recall for diverse queries, but increases indexing and storage complexity. It is ideal for dynamic retrieval but less practical for resource-constrained systems [25].

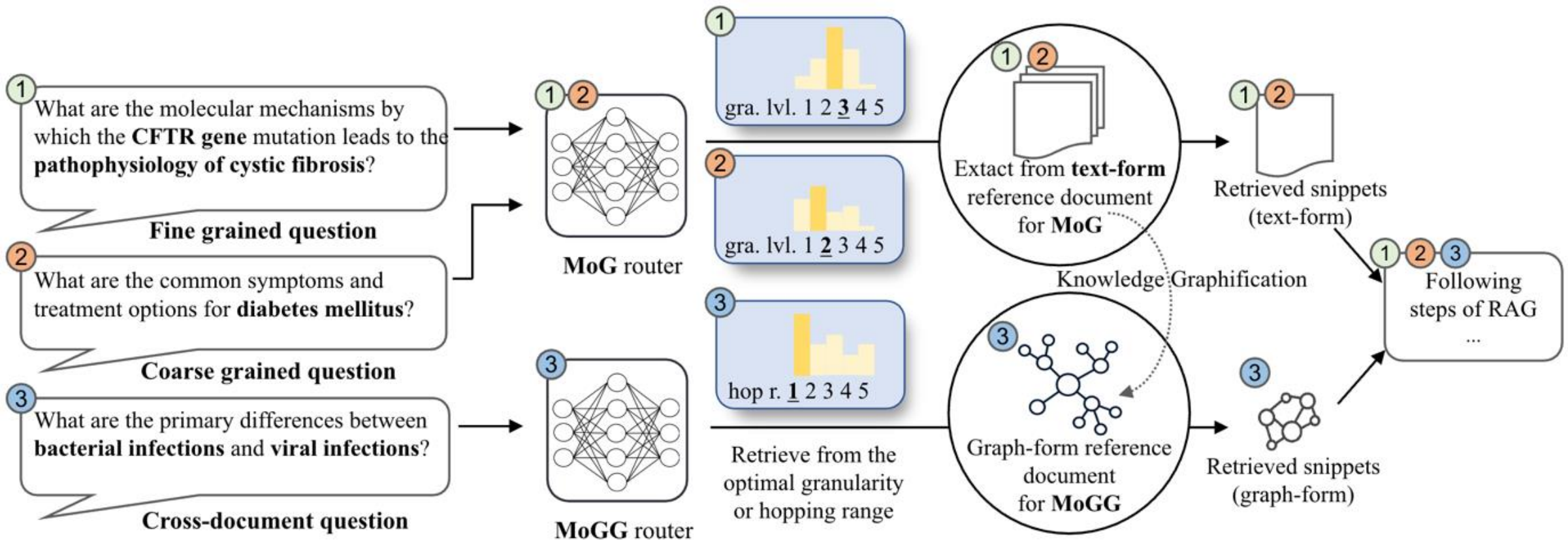

Figure 2: Mixture-of-Granularity (MoG) and Mixture-of-Granularity-and-Graph (MoGG) Retrieval Pipeline [25]

### 4.1.4 Comparative Analysis

Fixed-size chunking (non-overlapping, overlapping, sliding window, padding/clipping) is computationally efficient and scalable, ideal for large-scale indexing, but fragments context, reducing recall [16]. Semantically coherent chunking (sentence, paragraph, topic modeling, embedding, TextTiling) prioritizes meaning, enhancing precision and recall, but requires preprocessing or modeling, increasing costs [17]. Adaptive chunking (query-driven, LLM-informed, recursive, MoG) dynamically optimizes for query needs, balancing precision and recall, but introduces latency or resource demands [5].

1. **Precision vs. Recall**: Fixed-size methods favor precision for contained queries but lose recall due to fragmentation. Semantic methods improve both by preserving meaning, while adaptive methods fine-tune based on context.
2. **Computational Cost**: Fixed-size chunking is lightweight, followed by TextTiling and recursive methods. Topic modeling, embedding, and LLM-based methods are resource-intensive, with query-driven and MoG adding query-time costs.
3. **Use Cases**: Fixed-size suits batch processing, semantic methods excel in semantic retrieval, and adaptive methods are best for dynamic, query-specific tasks.

## 4.2 Image Chunking

Image chunking involves segmenting visual data into spatial regions for processing, indexing, or retrieval in multimodal AI systems. These regions serve as atomic units for tasks such as object detection, image retrieval, visual question answering, and multimodal Retrieval-Augmented Generation (RAG). Effective image chunking balances semantic richness, coverage, and computational efficiency, ensuring regions are meaningful and processable [31]. Image chunking strategies are categorized into five main methods: patch-based chunking (fixed grid), object detection-based chunking, region proposal-based chunking, dense captioning regions, and scene graph node chunking. Below, we provide a detailed analysis of each method, including mathematical formulations, practical considerations, and trade-offs in terms of precision, recall, and computational cost.

### 4.2.1 Patch-Based Chunking (Fixed Grid)

Patch-based chunking divides an image into a uniform grid of equal-sized patches, typically square or rectangular, without regard for semantic content. This method is computationally efficient and ensures complete coverage of the image, making it suitable for tasks like Vision Transformer (ViT) tokenization, where patches are treated as input tokens [12]. Each patch is processed independently, enabling parallel computation and scalability for large images. However, patches often split objects or semantic regions, leading to context fragmentation and low precision, as chunks may not align with meaningful visual units. The patch size $(K \times K)$ is a critical hyperparameter: smaller patches increase granularity but risk losing object-level context, while larger patches capture more context but reduce resolution. Edge cases include images smaller than the patch size, yielding a single patch, or non-divisible dimensions, requiring padding or clipping.

**Mathematical Formulation**: Let $(I \in R^{H \times W \times C})$ be an image with height (H), width (W), and (C) color channels. For a patch size $(K \times K)$, the set of patches $\left(\{P\} = \left\{P_{\{i,j\}} \mid i = 1, \dots, M, j = 1, \dots, N\right\}\right)$ is defined as:

[

$$P_{i,j} = I[(i-1)K{:}\,iK, (j-1)K{:}\,jK, :], \quad i = 1, \dots, M, \quad j = 1, \dots, N$$

]

where $(M = \lceil H/K \rceil), (N = \lceil W/K \rceil), and\left(P_{i,j} \in R^{K \times K \times C}\right)$. If $(H\backslash modK \neq 0) or (W\backslash modK \neq 0)$, the final patches are truncated or padded:

[

$$P_{M,j} = I[(M-1)K{:}\min(MK, H), (j-1)K{:}\min(jK, W), :], \quad \text{similarly for } P_{i,N}$$

]

Padding appends zeros to reach $(K \times K)$, while clipping retains available pixels. The total number of patches is$(M \cdot N)$.

**Practical Considerations**: Tools like PyTorch or TensorFlow support patch-based chunking for ViT preprocessing, splitting images into grids for tokenization [12]. The choice of (K) (e.g., 16 for ViT) depends on the model and task: smaller (K) suits fine-grained analysis, while larger (K) reduces computational load. The method is robust for high-resolution images but struggles with tasks requiring semantic alignment, such as object retrieval.

**Trade-offs**: Patch-based chunking is computationally efficient and scalable, with $\left(O(HW/K^2)\right)$ patches, ensuring high recall by covering the entire image. However, it sacrifices precision due to semantic fragmentation, making it suitable for tasks like image classification but less effective for object-specific retrieval [31].

## 4.2.2 Object Detection-Based Chunking

Object detection-based chunking uses pre-trained models (e.g., YOLO, Faster R-CNN) to identify objects and extract bounding boxes as chunks. Each chunk corresponds to a detected object, ensuring semantic relevance and high precision, as chunks align with meaningful visual entities [31]. This is ideal for tasks like visual question answering or object-centric retrieval, where queries target specific objects. However, the method depends on the detector's accuracy and coverage: missed objects reduce recall, and overlapping or nested boxes may introduce redundancy. The number of chunks varies with the image's content, complicating processing for images with many or few objects. Edge cases include images with no detected objects, producing no chunks, or crowded scenes, generating excessive chunks.

**Mathematical Formulation**: Let $(I \in R^{H\times W\times C})$ be an image, and let a detector output a set of bounding boxes $(B = B_1,\ B_2,\ \ldots,\ B_N)$,here $\left(B_i = (x_i, y_i, w_i, h_i)\right)$ defines the top-left corner $((x_i, y_i))$, width $(w_i)$, and height $(h_i)$. The set of chunks $(\mathcal{C} = C_1, C_2, \ldots, C_N)$ is:
[

$$C_i = I[x_i{:}\, x_i + w_i, y_i{:}\, y_i + h_i, :], \quad i = 1, \ldots, N$$

]
where $(C_i \in \{R\}^{\{h_i \times w_i \times C\}})$, and coordinates are clamped to $([0, H] \times [0, W])$. If no objects are detected, $(\mathcal{C} = \emptyset)$.

**Practical Considerations**: Frameworks like Detectron or YOLO provide robust object detection for chunking [31]. The choice of model affects chunk quality: models trained on diverse datasets (e.g., COCO) improve coverage. Post-processing (e.g., non-maximum suppression) reduces redundant boxes. The method is effective for images with distinct objects but struggles with abstract or texture-based images.

**Trade-offs**: Object detection-based chunking maximizes precision by focusing on semantic units but risks low recall if objects are missed. It is computationally intensive, with costs tied to detector complexity, but excels in object-centric tasks [31].

## 4.2.3 Region Proposal-Based Chunking

Region proposal-based chunking generates candidate regions using algorithms like Selective Search or models like OWL-ViT, which are then filtered to form chunks. This method maximizes recall by producing a large set of potential regions, capturing objects, parts, or background areas . It is suitable for tasks requiring comprehensive coverage, such as scene understanding or multimodal RAG. However, the large number of proposals introduces redundancy, reducing precision, as many regions may be irrelevant or overlapping. Filtering (e.g., via confidence scores or classifiers) mitigates this but increases computational cost. The number of chunks varies with the proposal algorithm and image complexity. Edge cases arise when images contain only a few distinguishable regions resulting in sparse or under-segmented chunks or, conversely, when complex scenes produce an excessive number of region proposals, leading to over-segmentation.

**Mathematical Formulation**: Let $(I \in ⅆo\{R\}\text{^}\{H \times W \times C\})$ be an image, and let a region proposal algorithm output a set of regions $(\mathcal{R} = R_1, R_2, \ldots, R_M)$, where $\left(R_i = (x_i, y_i, w_i, h_i)\right)$. A filtering function $(f: \mathcal{R} \rightarrow 0,1)$ (e.g., confidence threshold) selects chunks:

[

$$\mathcal{C} = C_i \mid C_i = I[x_i : x_i + w_i, y_i : y_i + h_i, :], f(R_i) = 1$$

]

The number of chunks $(|\mathcal{C}| \leq M)$ depends on (f). For example, if $(f(R_i) = 1)$when confidence $(\sigma_i > \tau)$, then:

[

$$\mathcal{C} = I[x_i : x_i + w_i, y_i : y_i + h_i, :] \mid \sigma_i > \tau$$

]

**Practical Considerations:** Tools such as Selective Search and OWL-ViT are commonly used to generate region proposals. OWL-ViT's open-vocabulary detection capability enhances flexibility by allowing the identification of a wide range of object categories without predefined labels. However, effective use of these tools requires careful tuning of filtering thresholds to balance precision and recall. While this method demonstrates strong performance in complex visual scenes, it remains computationally intensive due to the overhead of proposal generation and subsequent filtering.**Trade-offs:** Region proposal-based chunking offers high recall, as it captures a broad range of potential regions, but this comes at the cost of reduced precision because of redundant or overlapping segments. Despite its computational expense, it is particularly suitable for tasks that demand comprehensive scene coverage, such as visual question answering, object discovery, or scene understanding.

### 4.2.4 Dense Captioning Regions

Dense captioning identifies multiple regions within an image and associates each with a corresponding textual description, creating multimodal chunks that seamlessly combine visual and semantic information. Each chunk—comprising a region and its caption enhances interpretability and retrieval effectiveness in multimodal applications such as image-text retrieval, visual question answering, and content-based search [32]. This method balances precision and recall by focusing on semantically meaningful regions while providing contextual metadata via captions. However, it relies on the quality of the captioning model, and generating captions for many regions is computationally intensive. The number of chunks varies with image complexity and model granularity. Edge cases include images with few capturable regions, producing sparse chunks, or highly detailed images, generating numerous chunks.

**Mathematical Formulation**: Let $(I \in R^{H\times W\times C})$ be an image, and let a dense captioning model output a set of region-caption pairs $\big(\mathcal{D} = (R_1, T_1), (R_2, T_2), \dots, (R_N, T_N)\big)$, where $\big(R_i = (x_i, y_i, w_i, h_i)\big)$ and $(T_i)$ is a textual caption. The set of chunks $(\mathcal{C} = C_1, C_2, \dots, C_N)$ is:
[
$$C_i = (I[x_i{:}\,x_i + w_i, y_i{:}\,y_i + h_i, :], T_i), \quad i = 1, \dots, N$$
]
where $\big(C_i = (V_i, T_i)\big)$, with $(V_i \in R^{h_i\times w_i\times C})$as the visual chunk and $(T_i)$ as the textual metadata. If no regions are generated, $(\mathcal{C} = \emptyset)$.

**Practical Considerations**: Models like DenseCap or those trained on Visual Genome support dense captioning [32]. The choice of model affects caption quality and region granularity. Post-processing to filter low-confidence pairs improves precision. The method is effective for multimodal tasks but requires significant computational resources.

**Trade-offs**: Dense captioning balances precision and recall with rich, interpretable chunks but is computationally intensive. It excels in multimodal retrieval but struggles with resource-constrained settings [32].

### 4.2.5 Scene Graph Node Chunking

Scene graph node chunking uses a scene graph, where nodes represent objects and edges represent relationships, to define chunks. Each node's bounding box forms a visual chunk, augmented with relational metadata (e.g., object labels, edge attributes), enhancing semantic richness. This is ideal for tasks requiring contextual understanding, such as visual reasoning or multimodal RAG, where relationships between objects are critical. However, generating scene graphs is computationally intensive, requiring object detection, relation prediction, and graph construction. The method depends on the graph's quality, with errors in detection or relation inference reducing chunk accuracy. Edge cases include images with no objects, producing no chunks, or complex scenes, generating dense graphs with many chunks.

**Mathematical Formulation**: Let $(I \in R^{H\times W\times C})$ be an image, and let a scene graph $\big(G = (\mathcal{V}, \mathcal{E})\big)$ have nodes $(\mathcal{V} = V_1, V_2, \dots, V_N)$, where $\big(V_i = (B_i, L_i)\big)$ includes bounding box $\big(B_i = (x_i, y_i, w_i, h_i)\big)$ and label $(L_i)$, and edges $(\mathcal{E})$ denote relationships. The set of chunks

$(\mathcal{C} = C_1, C_2, \ldots, C_N)$ is:
[

$$C_i = (I[x_i : x_i + w_i, y_i : y_i + h_i, :], L_i, \mathcal{E}_i), \quad i = 1, \ldots, N$$

]
where $(\mathcal{E}_i \subseteq \mathcal{E})$ includes edges involving $(V_i)$, and the visual chunk is $(\in R^{h_i \times w_i \times C})$. If $(\nexists \mathrm{c}\{V\} = \emptyset), then\ (\nexists \mathrm{c}\{C\} = \emptyset)$.

**Practical Considerations**: Frameworks like Visual Genome or scene graph generation models support this method [30]. The choice of model affects graph quality, with robust detectors improving node accuracy. The method is effective for relational tasks but requires significant preprocessing and computational resources.

**Trade-offs**: Scene graph node chunking maximizes semantic richness, enhancing precision and recall for relational tasks, but is computationally expensive. It is ideal for complex visual reasoning but impractical for simple or resource-constrained applications [30].

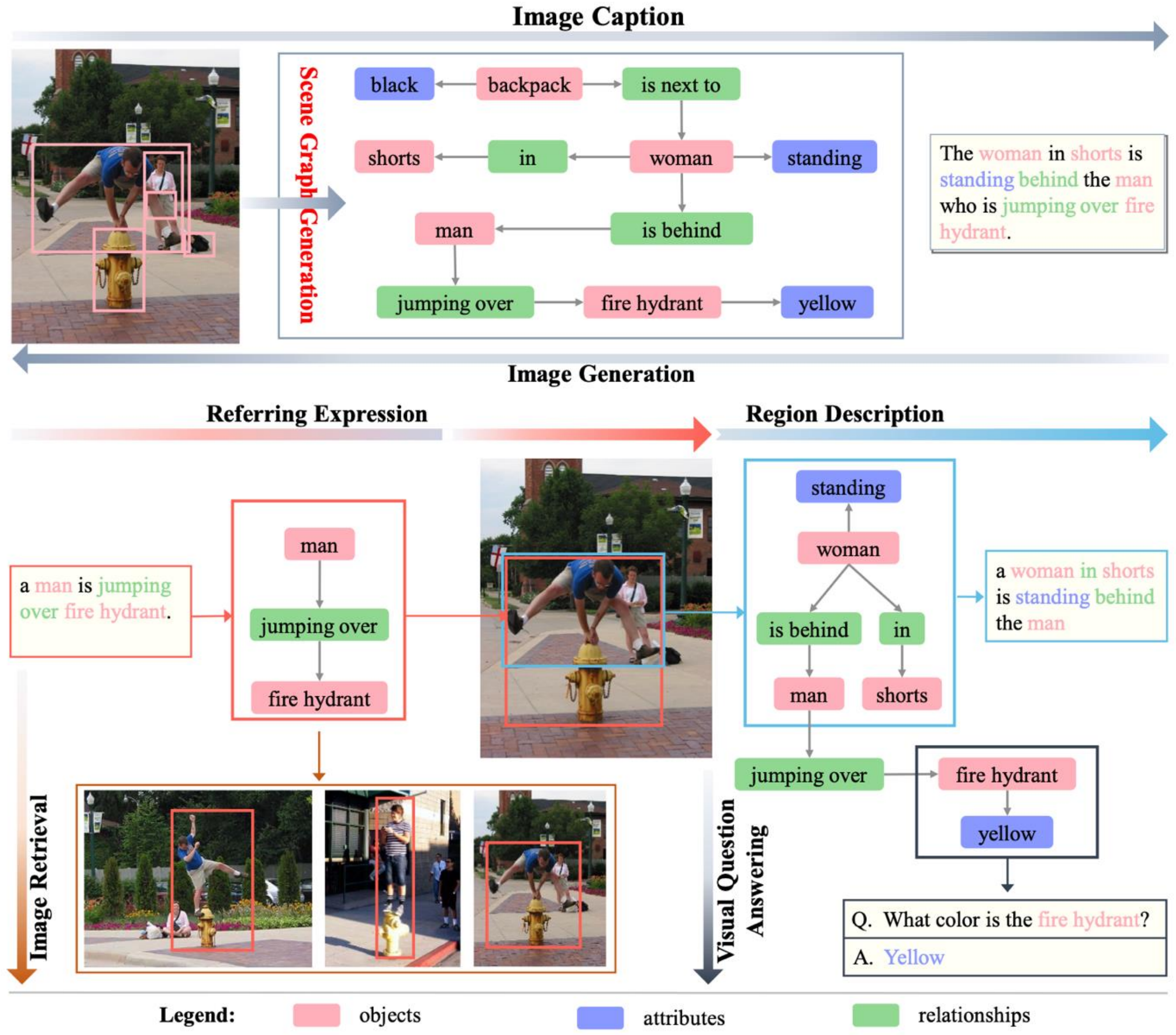

Figure 3: Scene Graph Node Chunking Overview
Illustrates how an input image is converted into a scene graph, where each node's bounding box and relational edges form individual visual semantic chunks and then used for tasks like captioning, referring expressions, retrieval, and VQA [30]

### 4.2.6 Comparative Analysis

Patch-based chunking is efficient and ensures high recall but sacrifices precision due to semantic fragmentation, making it suitable for global image processing [12]. Object detection-based chunking maximizes precision by focusing on semantic units but risks low recall if objects are missed, ideal for object-centric tasks [31]. Region proposal-based chunking prioritizes recall but introduces redundancy, fitting tasks requiring broad coverage. Dense captioning balances precision and recall with multimodal chunks, excelling in image-text retrieval but requiring heavy computation [32]. Scene graph node chunking offers rich semantics for relational tasks but is resource-intensive [30].

1. **Precision vs. Recall**: Patch-based chunking favors recall, object detection and dense captioning prioritize precision, region proposals maximize recall, and scene graphs balance both for relational queries.
2. **Computational Cost**: Patch-based chunking is lightweight, followed by object detection. Region proposals, dense captioning, and scene graphs are increasingly resource-intensive.
3. **Use Cases**: Patch-based suits classification, object detection fits retrieval, region proposals serve scene understanding, dense captioning excels in multimodal tasks, and scene graphs are best for relational reasoning [31].

## 4.3 Audio Chunking

Audio chunking segments audio data into discrete temporal segments for processing, indexing, or retrieval in multimodal AI systems. These segments are critical for tasks such as speech recognition, audio retrieval, sound event detection, and multimodal Retrieval-Augmented Generation (RAG). Effective audio chunking balances semantic coherence, temporal coverage, and computational efficiency, ensuring segments are meaningful and processable. Audio chunking strategies are categorized into four main methods: fixed-length chunking, silence-based chunking, speaker-based chunking, and content-based chunking. Below, we provide a detailed analysis of each method, including mathematically rigorous equations, practical considerations, and trade-offs in precision, recall, and computational cost.

### 4.3.1 Fixed-Length Chunking

Fixed-length chunking divides an audio signal into uniform temporal segments of a specified duration, typically measured in seconds or samples. This approach is computationally efficient and guarantees full coverage of the audio stream, making it well-suited for tasks such as automatic speech recognition (ASR) and audio feature extraction, where consistent segment sizes simplify downstream processing. However, it overlooks semantic and structural boundaries, such as pauses, speaker transitions, or sound events often resulting in fragmented context and reduced precision when meaningful audio cues are split across segments. The chunk duration (T) serves as a crucial hyperparameter: shorter chunks offer finer granularity but risk losing contextual continuity, while longer chunks capture broader context but may include irrelevant or overlapping information. Edge cases include audio shorter than (T), producing a single chunk, and non-divisible durations that necessitate padding or clipping to maintain uniform length.

**Practical Considerations:** Tools such as Librosa and PyTorch Audio support fixed-length chunking as part of the preprocessing pipeline. The optimal value of (T) commonly between 2–10 seconds for ASR tasks depends on the application. Shorter durations are preferred for fine-grained analysis, whereas longer durations reduce computational overhead in large-scale audio processing. While the method performs robustly for continuous audio streams, it is less effective for tasks that demand semantic or event-level alignment, such as dialogue segmentation or speaker diarization.

**Trade-offs:** Fixed-length chunking ensures high recall by covering the entire signal, with a computational complexity proportional to $(O(N/K))$. However, this comes at the cost of precision, as uniform segmentation may misalign with semantic events. Consequently, it is most appropriate for global audio tasks such as embedding generation or feature extraction—where efficiency and scalability outweigh the need for precise contextual segmentation.

### 4.3.2 Silence-Based Chunking

Silence-based chunking segments audio by detecting periods of silence, using energy thresholds or voice activity detection (VAD) to place boundaries. Each chunk typically corresponds to a speech utterance or non-silent event, ensuring semantic coherence and high precision for tasks like speech recognition or audio retrieval, where queries target spoken content [33]. However, the method's recall depends on the VAD's accuracy: missed silences merge distinct events, while false positives fragment utterances. It is sensitive to noise, requiring robust preprocessing. The number of chunks varies with audio content, complicating processing for noisy or continuous audio. Edge cases include silent audio, producing no chunks, or highly noisy audio, yielding fragmented segments.

**Mathematical Formulation**: Let $(A \in R^N)$ be an audio signal, and let a VAD algorithm output silence boundaries$(b_0, b_1, \ldots, b_L)$, where $(b_0 = 0)$, $(b_L = N)$, and $(b_i)$ are sample indices of silence transitions. The set of chunks $(\{C\} = \{\ C_i \mid i = 1, \ldots, L\ \})$ is:

[

$$C_i = A[b_{i-1} : b_i], \quad i = 1, \ldots, L,$$

]

where ($C_i \in R^{b_i - b_{i-1}}$), and (L) is the number of non-silent segments. The VAD function ($v: R^N \to 0,1^N$) labels samples as silence (0) or non-silence (1), with boundaries at transitions:
[
$b_i = \min j > b_{i-1} \mid v(A[j]) \neq v(A[j-1])$ or $j = N.$
]
If ($v(A) = 0$) for all samples, ($\mathcal{C} = \emptyset$).

**Practical Considerations**: Tools like WebRTC VAD or PyAnnote provide robust silence detection [33]. Threshold tuning (e.g., energy or probability) balances sensitivity to silence. The method excels for dialogue or podcast audio but struggles with music or ambient noise, requiring noise reduction preprocessing.

**Trade-offs**: Silence-based chunking maximizes precision for speech-centric tasks but risks low recall if silences are misdetected. It is moderately computationally intensive, depending on VAD complexity, and is ideal for utterance-based retrieval [33].

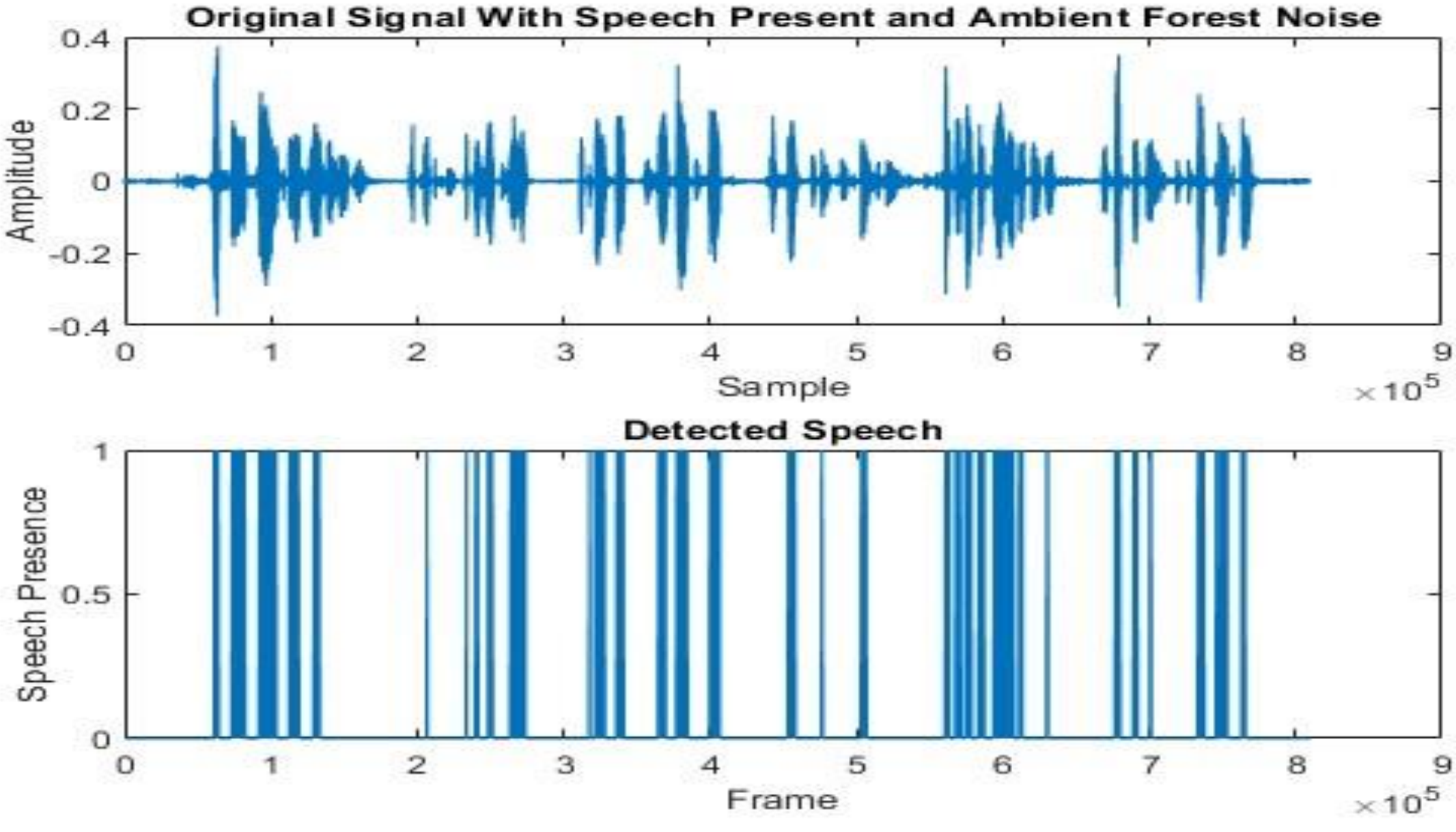

Figure 4: This graphic depicts the mixed audio signal along with the result of the detected speech in a binary format on the specified frame [33]

### 4.3.3 Speaker-Based Chunking

Speaker-based chunking segments audio by identifying speaker changes, using speaker diarization to assign segments to distinct speakers. Each chunk represents a speaker's continuous speech, ensuring semantic coherence for tasks like meeting transcription or dialogue analysis, where speaker context is critical [34]. The method's precision depends on diarization accuracy: errors in speaker identification merge or split segments incorrectly, reducing recall. It is computationally intensive, requiring clustering or embedding-based diarization. The number of chunks varies with speaker transitions, complicating processing for single-speaker or multi-speaker audio. Edge cases include single-speaker audio, producing one chunk, or overlapping speech, yielding ambiguous segments.

**Mathematical Formulation**: Let ($A \in R^N$) be an audio signal. A diarization algorithm outputs speaker boundaries ($b_0, b_1, \ldots, b_L$), where($b_0 = 0$), ($b_L = N$), and ($b_i$) are sample indices of speaker transitions, with labels ( $s_i \in \mathcal{S} \mid i = 1, \ldots, L$ ), where ($\mathcal{S}$) is the set of speakers. The set of chunks ($\{C\} = \{ C_i \mid i = 1, \ldots, L \}$) is:

[

$$C_i = (A[b_{i-1}: b_i], s_i), \quad i = 1, \ldots, L,$$

]

where $\left(C_i = (A_i, s_i)\right)$, with ($A_i \in R^{b_i - b_{i-1}}$) as the audio segment and ($s_i$) as the speaker label. If (L = 1), ($\mathcal{C}$) contains one chunk.

**Practical Considerations**: Tools like PyAnnote or NeMo support diarization, using embeddings (e.g., x-vectors) for clustering [34]. The method excels for multi-speaker audio like meetings but requires robust preprocessing for overlapping speech or noise. Tuning clustering parameters ensures accurate speaker separation.

**Trade-offs**: Speaker-based chunking maximizes precision for speaker-specific tasks but risks low recall due to diarization errors. It is computationally expensive but ideal for dialogue or transcription tasks requiring speaker context [34].

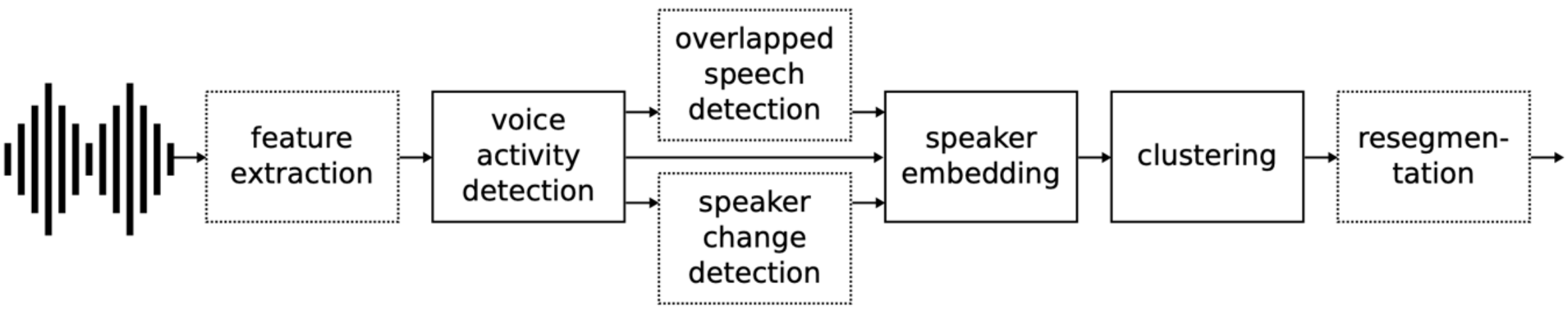

Figure 5: pyannote.audio provides a collection of modules that can be jointly optimized to build a speaker diarization pipeline. [34]

### 4.3.4 Content-Based Chunking

Content-based chunking segments audio based on semantic or acoustic content, such as sound events, topics, or speech patterns, using techniques like acoustic modeling, topic detection, or neural embeddings. This method ensures high semantic coherence, making it suitable for tasks like audio event detection, topic-based retrieval, or multimodal RAG, where queries target specific content. It is highly flexible, adapting to audio characteristics, but computationally intensive, requiring trained models or embeddings. The method's precision and recall depend on model accuracy, with errors fragmenting or merging segments. The number of chunks varies with content complexity. Edge cases include uniform audio (e.g., continuous noise), producing few chunks, or highly varied audio, yielding many segments.

**Mathematical Formulation**: Let ($A \in R^N$)be an audio signal. A content detection model outputs boundaries ($b_0, b_1, \ldots, b_L$), where ($b_0 = 0$), ($b_L = N$), and ($b_i$) are sample indices of content shifts, with labels ( $l_i \in \mathcal{L} \mid i = 1, \ldots, L$ ), where ($\mathcal{L}$) is the set of content types (e.g., topics, events). The set of chunks ($\{C\} = \{ C_i \mid i = 1, \ldots, L \}$) is:

[

$$C_i = (A[b_{i-1}: b_i], l_i), \quad i = 1, \ldots, L,$$

]

where $\left(C_i = (A_i, l_i)\right)$, with ($A_i \in R^{b_i - b_{i-1}}$) and ($l_i$) as the content label. Boundaries are determined by a similarity function (e.g., cosine similarity of embeddings ($e_i \in \{R\}^d$)):

[

$$b_i = \min j > b_{i-1} \frac{1}{e_j \cdot e_{j+1} \& |e_j| 2 \, | \, ej + 1|_2 < \tau \text{ or } j = N,}$$

]

where($\tau \in [0,1]$) is a similarity threshold. If no shifts are detected, (L = 1).

**Practical Considerations**: Tools like Hugging Face's audio transformers or Librosa support content-based chunking via embeddings or acoustic features. Models trained on datasets like AudioSet improve event detection. Threshold tuning balances granularity. The method is robust for diverse audio but requires significant computational resources.

**Trade-offs**: Content-based chunking maximizes precision and recall for content-specific tasks but is computationally intensive. It excels in semantic retrieval but is impractical for resource-constrained settings.

### 4.3.5 Comparative Analysis

Fixed-length chunking is efficient and ensures high recall but fragments context, reducing precision, making it suitable for global processing like ASR. Silence-based chunking prioritizes precision for speech-centric tasks but risks low recall due to VAD errors, ideal for utterance-based retrieval [33]. Speaker-based chunking maximizes precision for speaker-specific tasks but is computationally expensive, fitting dialogue analysis [34]. Content-based chunking offers high semantic coherence, balancing precision and recall, but requires heavy computation, excelling in event or topic-based retrieval.

1. **Precision vs. Recall**: Fixed-length chunking favors recall, silence-based and speaker-based prioritize precision, and content-based balances both for semantic queries.
2. **Computational Cost**: Fixed-length chunking is lightweight, followed by silence-based. Speaker-based and content-based are increasingly resource-intensive.
3. **Use Cases**: Fixed-length suits feature extraction, silence-based fits speech retrieval, speaker-based excels in transcription, and content-based is best for semantic tasks.

# 4.4 Video Chunking

Video chunking segments video data into discrete temporal or spatiotemporal segments for processing, indexing, or retrieval in multimodal AI systems. These segments are critical for tasks such as video retrieval, action recognition, video summarization, and multimodal RAG. Effective video chunking balances semantic coherence, temporal coverage, and computational efficiency, ensuring segments capture meaningful events or actions while remaining processable. Video chunking strategies are categorized into four main methods: fixed-length temporal chunking, shot-based chunking, action-based chunking, and semantic-based chunking. Below, we provide a detailed analysis of each method, including mathematically rigorous equations, practical considerations, and trade-offs in precision, recall, and computational cost.

## 4.4.1 Fixed-Length Temporal Chunking

Fixed-length temporal chunking divides a video into uniform time-based segments of a specified duration, typically measured in seconds or frames. This approach is computationally efficient and ensures complete temporal coverage, making it well-suited for tasks such as video feature extraction or preprocessing for 3D convolutional neural networks (CNNs). However, it disregards semantic and structural boundaries—such as scene transitions, camera cuts, or distinct actions—potentially fragmenting meaningful events and reducing contextual coherence. The segment duration (T) is a crucial hyperparameter: shorter segments provide finer temporal resolution but risk losing broader context, while longer segments capture more information at the expense of introducing irrelevant content. Edge cases include short videos (shorter than T), which collapse into a single segment, and non-divisible durations, which require padding or clipping to achieve uniform segmentation.

**Practical Considerations:** Libraries such as *OpenCV* and *FFmpeg* support fixed-length temporal segmentation, enabling efficient preprocessing workflows. The optimal choice of T (typically between 5–30 seconds) depends on the downstream application—shorter segments are advantageous for fine-grained event analysis, whereas longer segments reduce computational cost in large-scale processing. This method performs reliably for continuous video streams but struggles in scenarios requiring semantic or event-level alignment, such as scene boundary detection or action segmentation.

**Trade-offs:** Fixed-length temporal chunking achieves high recall by ensuring full video coverage, producing approximately $O(F/K)$ segments (where F is the total number of frames and K the segment length). However, it sacrifices precision by ignoring semantic structure, which can lead to context fragmentation. As a result, this approach is best suited for global video analysis tasks, such as embedding generation or feature extraction, where scalability and efficiency take precedence over semantic fidelity.

### 4.4.2 Shot-Based Chunking

Shot-based chunking divides video content by detecting shot boundaries, where each shot represents a continuous sequence of frames captured by a single camera take. These boundaries are typically identified using visual cues such as abrupt frame differences, histogram variations, or motion discontinuities ensuring that each chunk corresponds to a visually coherent unit [35]. This method is particularly effective for video summarization, retrieval, and film analysis, where queries often target individual shots or scenes. However, its accuracy depends heavily on the shot detection algorithm: missed boundaries can merge distinct shots, while false positives may fragment them. Gradual transitions, such as fades or dissolves, further complicate detection and require robust algorithms capable of handling subtle visual changes. The number of resulting chunks varies with shot frequency, posing challenges for videos that are either highly static (few shots) or rapidly edited (many short shots). Edge cases include single-shot videos, which yield a single segment, and rapid-cut sequences, which produce numerous short chunks that increase processing overhead.

**Practical Considerations:** Tools such as *PySceneDetect* and *OpenCV* offer reliable shot detection based on frame-difference metrics, color histograms, or motion analysis [35]. Careful tuning of detection thresholds is essential to balance sensitivity to abrupt cuts against robustness to gradual transitions. The approach performs best on professionally edited content such as films, news segments, or advertisements, but may require noise reduction or stabilization preprocessing when applied to raw or continuous footage.

**Trade-offs:** Shot-based chunking achieves high precision by aligning segments with natural scene boundaries, making it ideal for structured video analysis tasks like summarization or retrieval. However, it can suffer from reduced recall due to imperfect boundary detection and is moderately computationally intensive, depending on the complexity of the detection method used. Overall, this approach offers a strong balance between semantic coherence and processing efficiency for edited or narrative-driven video content [35].

### 4.4.3 Action-Based Chunking

Action-based chunking segments video content by detecting boundaries between distinct actions or events, leveraging models trained to recognize specific activities (e.g., walking, jumping) or general event changes. Each chunk corresponds to a semantically coherent action, making this method particularly suitable for tasks such as action recognition or event-based retrieval [37]. The method's precision is contingent on the accuracy of the underlying action detection model: misclassifications can merge distinct actions or split single actions across multiple chunks, thereby reducing recall. Action-based chunking is computationally intensive, typically requiring pretrained models or action embeddings. The number of generated chunks varies with the frequency of actions, which complicates processing for videos with very sparse or very dense activity. Edge cases include videos lacking recognizable actions, yielding few chunks, or sequences with continuous activity, resulting in overly large segments.

**Mathematical Formulation**: Let $(V \in R^{F\times H\times W\times C})$ be a video. An action detection model outputs boundaries $(b_0, b_1, \dots, b_L)$, where $(b_0 = 0), (b_L = F), and (b_i)$ are frame indices of action transitions, with labels $( a_i \in \mathcal{A} \mid i = 1, \dots, L ), where (\mathcal{A})$ is the set of action types. The set of chunks $(\{C\} = \{ C_i \mid i = 1, \dots, L \})$ is:
[

$$C_i = (V[b_{i-1}{:}\, b_i, :, :, :], a_i), \quad i = 1, \dots, L,$$

]
where $\big(C_i = (V_i, a_i)\big), with \big(V_i \in R^{(b_i - b_{i-1})\times H\times W\times C}\big) and (a_i)$ as the action label. Boundaries are determined by a classifier or embedding similarity:
[

$$b_i = \min t > b_{i-1} \frac{1}{e_t \cdot e_{t+1} \& |e_t|2 \,\big|\, et + 1|_2 < \tau \text{ or } t = F},$$

]
where $(e_t)$ are frame or clip embeddings, and $(\tau)$ is a similarity threshold. If no actions are detected, $(L = 1)$.

**Practical Considerations**: Models like I3D or SlowFast, trained on datasets like Kinetics, support action-based chunking [37]. Tools like PyTorch or Hugging Face provide pretrained models. Threshold tuning balances granularity. The method is robust for action-rich videos but requires significant computational resources and struggles with ambiguous or overlapping actions.

**Trade-offs**: Action-based chunking maximizes precision for action-specific tasks but risks low recall due to detection errors. It is computationally expensive but ideal for event-based retrieval or recognition [37].

### 4.4.4 Semantic-Based Chunking

Semantic-based chunking segments a video based on narrative or thematic shifts, using multimodal cues (e.g., visual, audio, or transcript features) to identify coherent segments. Models like CLIP or video captioning systems analyze content to detect changes in context, ensuring high semantic coherence [36]. This method is suitable for tasks like video summarization, thematic retrieval, or multimodal RAG, where queries target narrative units. It is highly flexible but computationally intensive, requiring multimodal models or embeddings. The method's precision and recall depend on model accuracy, with errors fragmenting or merging segments. The number of chunks varies with content complexity. Edge cases include uniform videos (e.g., single scene), producing few chunks, or highly varied videos, yielding many segments.

**Mathematical Formulation**: Let $(V \in R^{F\times H\times W\times C})$be a video, with associated audio $(A \in R^{N})$ and transcript (T ). A semantic detection model outputs boundaries $(b_0, b_1, \dots, b_L), where(b_0 = 0), (b_L = F), and(b_i)$ are frame indices of semantic shifts, with labels $( s_i \in \mathcal{S} \mid i = 1, \dots, L ), where(\mathcal{S})$ is the set of semantic themes. The set of chunks $(\{C\} = \{ C_i \mid i = 1, \dots, L \})$ is:
[

$$C_i = \left(V[\quad _{i-1}: \quad _{i}, :, :, : ], A\left[\left\lfloor b_{i-1} \cdot \frac{N}{F}\right\rfloor : \left\lfloor b_i \cdot \frac{N}{F}\right\rfloor\right], T_i, s_i\right), \quad i = 1, \dots, L,$$

]
where $\left(C_i = (V_i, A_i, T_i, s_i)\right), with\left(V_i \in R^{(b_i - b_{i-\mathbb{1}})\times H\times W\times C}\right), \left(A_i \in R^{\left\lfloor (b_i - b_{i-\mathbb{1}})\cdot \frac{N}{F}\right\rfloor}\right), (T_i)$ as the transcript segment, and $(s_i)$ as the theme. Boundaries are determined by multimodal embedding similarity:
[

$$b_i = \min t > b_{i-1} \frac{1}{e_t \cdot e_{t+1} \& |e_t| 2 \left| et + 1 \right|_2 < \tau \text{ or } t = F,}$$

]
where $(e_t)$ are multimodal embeddings (e.g., CLIP). If no shifts are detected, (L = 1).

**Practical Considerations**: Tools like CLIP or VideoCLIP support semantic chunking by analyzing visual and textual cues [36]. Preprocessing (e.g., transcript generation with Whisper) enhances accuracy. The method is robust for narrative videos but requires significant computational resources and struggles with abstract or non-narrative content.

**Trade-offs**: Semantic-based chunking maximizes precision and recall for thematic tasks but is computationally intensive. It excels in multimodal retrieval but is impractical for resource-constrained settings [36].

### 4.4.5 Comparative Analysis

Fixed-length chunking is efficient and ensures high recall but fragments context, reducing precision, making it suitable for global processing like feature extraction. Shot-based chunking prioritizes precision for shot-specific tasks but risks low recall due to detection errors, ideal for summarization. Action-based chunking maximizes precision for action-specific tasks but is computationally expensive, fitting event-based retrieval [37]. Semantic-based chunking offers high semantic coherence, balancing precision and recall, but requires heavy computation, excelling in thematic tasks [36].

1. **Precision vs. Recall**: Fixed-length chunking favors recall, shot-based and action-based prioritize precision, and semantic-based balances both for thematic queries.
2. **Computational Cost**: Fixed-length chunking is lightweight, followed by shot-based. Action-based and semantic-based are increasingly resource-intensive.
3. **Use Cases**: Fixed-length suits feature extraction, shot-based fits summarization, action-based excels in event retrieval, and semantic-based is best for thematic tasks.

# 5. Cross-Modal Chunking

Cross-modal chunking integrates data from multiple modalities (e.g., text, images, audio, video) into unified, semantically coherent units for processing, indexing, or retrieval in multimodal AI systems. These units are critical for tasks like multimodal RAG, cross-modal search, or integrated reasoning, where alignment across modalities enhances performance. Effective cross-modal chunking ensures semantic consistency, modality alignment, and processability, balancing complexity and coherence. Cross-modal chunking strategies are categorized into three main methods: text-image alignment, audio-transcript synchronization, and joint multimodal embedding-based chunking. Below, we provide a detailed analysis of each method, including mathematical formulations, practical considerations, and trade-offs.

## 5.1 Text-Image Alignment

Text-image alignment chunks documents by grouping images with their corresponding textual descriptions, using structural or semantic cues (e.g., layout, captions, or proximity in documents). Each chunk represents a text–image pair, maintaining semantic coherence between visual and linguistic content. This structure is particularly effective for multimodal tasks such as document retrieval or visual question answering, where contextual alignment between modalities is essential. The method is computationally lightweight, as it leverages existing document structures (e.g., HTML tags, PDF layouts) to guide chunk formation. However, its precision depends heavily on the reliability of these structural cues—missing or ambiguous associations can weaken alignment and lower recall. The number of resulting chunks typically scales with document complexity. Edge cases include text-only documents, which yield purely textual chunks, and image-dense materials, which produce a large number of texts–image pairs.

**Mathematical Formulation**: Let $\left(D = (T, \mathcal{J})\right)$ be a document with text $(T = [t_1, t_2, \dots, t_N])$ and images $(\mathcal{J} = I_1, I_2, \dots, I_M)$, where $(I_i \in \{R\}^{\{H_i \times W_i \times C\}}\}))$ . An alignment function ($f: \mathcal{I} \rightarrow \mathcal{T}$) maps each image to a text segment $\left(T_j \subseteq T\right)$. The set of chunks ($\{C\} = \{\ C_i \mid i\ =\ 1, \dots, L\ \}$) is:
[

$$C_i = (T_i, I_i), \quad \text{where } T_i = f(I_i), \quad i = 1, \dots, M,$$

]
plus text-only chunks for unmapped segments:
[

$$C_{M+j} = \left(T_j, \emptyset\right), \quad \text{for } T_j / \in f(I_i) \mid i = 1, \dots, M.$$

]
The total number of chunks ($L\ \geq M$) depends on text segments and unmapped text. The alignment function (f) may use layout cues (e.g., proximity in PDFs) or captions.

**Practical Considerations**: Tools like PDFMiner or BeautifulSoup extract text-image pairs from PDFs or HTML. Preprocessing to normalize layouts or detect captions enhances accuracy. The method is robust for structured documents (e.g., articles) but struggles with unstructured data or missing metadata.

**Trade-offs**: Text-image alignment is lightweight and precise for structured documents but risks low recall if alignments are missed. It is ideal for document-based multimodal tasks but less effective for complex or unstructured data.

## 5.2 Audio-Transcript Synchronization

Audio-transcript synchronization chunks audio data with its corresponding transcript, aligning speech segments with textual representations. Each chunk is an audio-transcript pair, ensuring semantic coherence for tasks like speech retrieval or multimodal RAG [38]. This method is precise, leveraging timestamps from automatic speech recognition (ASR), but its recall depends on ASR accuracy: errors in transcription or alignment fragment or merge segments. The method is moderately computationally intensive, requiring ASR and alignment processing. The number of chunks varies with speech complexity. Edge cases include audio with no speech, producing no chunks, or noisy audio, yielding misaligned segments.

**Mathematical Formulation**: $Let(A \in R^N)$ be an audio signal at sampling rate $(f_s)$, and let $(T = [t_1, t_2, \dots, t_M])$ be its transcript with timestamps $\left(\ (s_i, e_i) \mid i = 1, \dots, M\ \right), where\ (s_i, e_i)$ are start and end times in seconds. The set of chunks ($\{C\} = \{\ C_i \mid i\ =\ 1, \dots, M\ \}$) is:
[

$$C_i = (A[\lfloor s_i \cdot f_s \rfloor : \lfloor e_i \cdot f_s \rfloor], t_i), \quad i = 1, \dots, M,$$

]
where $\left(C_i = (A_i, t_i)\right), with\ \left(A_i \in R^{\lfloor (e_i - s_i) \cdot f_s \rfloor}\right) and (t_i)$ as the transcript segment. If no transcript is generated,($\mathcal{C} = \emptyset$).

**Practical Considerations**: Tools like Whisper or Kaldi provide ASR with timestamps for synchronization [38]. Post-processing to correct alignment errors improves precision. The method excels for speech-heavy audio (e.g., lectures) but struggles with noisy or non-speech audio, requiring preprocessing.

**Trade-offs**: Audio-transcript synchronization maximizes precision for speech-based tasks but risks low recall due to ASR errors. It is moderately computationally intensive and ideal for multimodal retrieval involving transcripts [38].

## 5.3 Joint Multimodal Embedding-Based Chunking

Joint multimodal embedding-based chunking segments data by clustering multimodal embeddings (e.g., from CLIP or VideoCLIP) that combine features from text, images, audio, or video. Each chunk is a multimodal unit, ensuring semantic coherence across modalities for tasks like cross-modal search or integrated reasoning . This method is highly flexible, adapting to unstructured data, but computationally intensive, requiring pretrained multimodal models. The method's precision and recall depend on embedding quality, with noisy embeddings reducing accuracy. The number of chunks varies with data complexity. Edge cases include unimodal data, reducing to modality-specific chunking, or highly varied data, yielding many chunks.

**Mathematical Formulation**: Let $(D = (T, \text{Ⅎc}\{I\}, A, V)$ be a multimodal document with text (T ), images $(\mathcal{I}), audio(A)$, and video (V ). A multimodal model generates embeddings$( e_i \in R^d \mid i = 1, \ldots, N )$ for segments (e.g., sentences, image regions, audio clips, video clips). Boundaries are placed where embedding similarity drops:
[

$$b_i = \min j > b_{i-1} \frac{1}{e_j \cdot e_{j+1} \& |e_j| 2 \mid ej + 1|2\} < \tau \text{ or } j = N,}$$

]
The set of chunks $(\{C\} = \{ C_i \mid i = 1, \ldots, L \})$is:
[

$$C_i = (T[bi - 1 : b_i], \mathcal{J}i, A[bi - 1 : b_i], V[b_{i-1} : b_i]), \quad i = 1, \ldots, L,$$

]
where $(\mathcal{J}_i \subseteq \mathcal{J})$ includes images within the segment, and boundaries are aligned across modalities. If no shifts are detected, (L = 1).

**Practical Considerations**: Tools like CLIP or VideoCLIP support multimodal embeddings . Preprocessing to segment modalities (e.g., text sentences, video shots) facilitates embedding generation. The method is robust for complex, unstructured data but requires significant computational resources.

**Trade-offs**: Joint multimodal chunking maximizes precision and recall for cross-modal tasks but is computationally intensive. It excels in integrated reasoning but is impractical for resource-constrained settings.

## 5.4 Comparative Analysis

Text-image alignment is lightweight and precise for structured documents but risks low recall if alignments are missed, ideal for document-based tasks. Audio-transcript synchronization maximizes precision for speech-based tasks but depends on ASR accuracy, fitting multimodal retrieval [38]. Joint multimodal chunking offers high semantic coherence, balancing precision and recall, but requires heavy computation, excelling in cross-modal search .

1. **Precision vs. Recall**: Text-image alignment and audio-transcript synchronization prioritize precision, while joint multimodal chunking balances both for complex queries.
2. **Computational Cost**: Text-image alignment is lightweight, audio-transcript synchronization is moderately intensive, and joint multimodal chunking is resource-heavy.
3. **Use Cases**: Text-image alignment suits document retrieval, audio-transcript synchronization fits speech retrieval, and joint multimodal chunking excels in cross-modal tasks.

# 6. Challenges, Future Directions, and Conclusion

Multimodal chunking is a cornerstone of modern AI systems, enabling the coherent and efficient processing of diverse data types, including text, images, audio, and video. As multimodal AI systems become integral to applications such as RAG, semantic search, and multimedia analytics, the design of effective chunking strategies has emerged as a critical research and engineering challenge. Despite significant advancements, several technical, practical, and ethical hurdles persist, limiting the scalability, robustness, and fairness of chunking methods. This chapter synthesizes the key challenges in multimodal chunking, proposes future research directions to address these issues, and concludes with a reflection on the broader implications for AI system design and deployment.

## 6.1 Challenges and Open Problems

The development of robust multimodal chunking strategies faces several interconnected challenges, each impacting the performance, scalability, and applicability of AI systems. Below, we detail seven critical challenges, supported by recent literature and practical insights.

### 1. Semantic Coherence Across Modalities

Ensuring that chunks preserve semantic meaning across modalities (e.g., aligning text with corresponding images or audio with transcripts) is a fundamental challenge. Misaligned or fragmented chunks can lead to false associations, reducing retrieval accuracy and generative coherence. For instance, in multimodal RAG, a text chunk describing an image may be paired incorrectly if layout cues are ambiguous, resulting in irrelevant responses. This issue is exacerbated in unstructured data, such as web pages or user-generated content, where explicit alignment signals (e.g., captions, timestamps) are often missing or noisy [14]. Current methods like CLIP-based embeddings struggle with fine-grained alignment in complex documents, highlighting the need for more robust cross-modal segmentation techniques .

**2. Scalability and Efficiency**

Advanced chunking methods, particularly those relying on neural models (e.g., embedding-based or LLM-informed segmentation), are computationally intensive, limiting their applicability to large-scale or real-time systems. For example, generating embeddings for high-resolution images or long videos requires significant computational resources, making it impractical for edge devices or low-latency applications. Fixed-size chunking is scalable but sacrifices semantic quality, while adaptive methods like Mixture-of-Granularity (MoG) increase indexing complexity by maintaining multiple chunk versions [25]. Balancing granularity with processing speed remains a critical trade-off, especially for enterprise-scale datasets with millions of multimodal documents.

**3. Robustness to Noise**

Noisy inputs, such as OCR errors in text, background noise in audio, or low-quality video frames, disrupt chunking accuracy. For instance, silence-based audio chunking using Voice Activity Detection (VAD) may fail in noisy environments, fragmenting utterances or merging distinct speech segments [33]. Similarly, object detection-based image chunking can miss objects in blurry images, reducing recall. Current preprocessing techniques, such as denoising or data augmentation, are insufficient for highly variable real-world data, necessitating error-tolerant chunking methods that maintain performance under adverse conditions. This challenge is particularly acute in domains like medical imaging or surveillance, where noise is prevalent.

**4. Generalization Across Domains**

Many chunking strategies are tailored to specific domains (e.g., legal documents, film analysis), limiting their adaptability to diverse applications. For example, paragraph-level text chunking assumes well-structured documents, which may not hold for social media posts or OCR outputs [21]. Similarly, shot-based video chunking is effective for edited films but struggles with unedited livestreams [35]. Domain-specific methods often require retraining or fine-tuning, increasing development costs and hindering generalization to new tasks or datasets. Developing domain-agnostic chunking frameworks that adapt to varying data characteristics remains an open problem.

**5. Evaluation Metrics**

The absence of standardized evaluation metrics for multimodal chunking limits comparative analyses and the systematic development of new methods. Existing metrics, such as precision and recall in retrieval tasks, are typically task-specific and insufficient for assessing cross-modal alignment or semantic coherence. For example, evaluating text-image alignment requires metrics that jointly consider both visual and textual relevance, yet such standards are rarely established [14]. Furthermore, the scarcity of benchmark datasets for multimodal chunking compounds the challenge, as most existing datasets focus on unimodal tasks or narrowly defined applications. Establishing standardized evaluation metrics and benchmark datasets is essential to drive progress and ensure reproducibility in research.

**6. Temporal and Spatial Dynamics**

Dynamic content such as changing video actions or shifting speakers in audio—creates major challenges for static chunking methods that rely on fixed boundaries. For example, semantic-based video chunking must capture subtle narrative changes that differ in length and complexity, requiring more adaptive temporal segmentation. Likewise, spatial variations in images, like overlapping objects, make object-based chunking difficult since simple bounding boxes often miss contextual relationships. Many current methods still depend on heuristic thresholds or pretrained models, which can struggle with such variability underscoring the need for real-time, adaptive chunking strategies that adjust to dynamic content.

**7. Ethical Considerations**

Pretrained models used for chunking like CLIP can carry inherent biases toward certain demographics or cultural contexts, which may influence chunking outcomes and affect fairness. For example, biased object detection might overrepresent specific objects, leading to skewed retrieval results in areas like surveillance or content moderation. Similarly, speaker-based audio chunking can struggle with underrepresented accents or dialects, resulting in unequal performance. To address these issues, it's essential to audit and mitigate biases in both training data and models, and to maintain transparency in how chunks are created and used. This is especially important in sensitive fields such as healthcare, law, and education.

## 6.2 Future Directions

Overcoming the challenges discussed above calls for ongoing innovation in both research and engineering. We outline eight promising directions for advancing multimodal chunking, guided by recent trends and insights from existing literature.

**1. Hybrid Chunking Frameworks**

Combining rule-based, statistical, and neural chunking approaches can leverage their individual strengths to improve both efficiency and semantic coherence. For instance, recursive text splitting can be paired with embedding-based segmentation to balance scalability and contextual precision [25]. Hybrid pipelines may first apply rule-based methods for broad segmentation, then use neural models for fine-tuning boundaries reducing computational load without sacrificing quality. Further research is needed to determine the most effective combinations and transitions between these methods for building robust, general-purpose chunking systems.

**2. Self-Supervised Learning for Chunking**

Self-supervised learning (SSL) provides a way to reduce reliance on labeled data by allowing models to learn boundaries directly from unlabeled multimodal datasets. Models like VideoCLIP show that pretraining on large-scale video-text pairs can identify semantic transitions without manual annotation [38]. Extending SSL to cross-modal alignment for example, linking images with text or audio with transcripts can further enhance generalization across domains. Future research should focus on developing SSL architectures specifically optimized for chunking, prioritizing lightweight and adaptable designs for a wide range of applications.

**3. Real-Time Chunking Methods**

Applications like streaming video analysis and interactive AI systems require lightweight, low-latency chunking algorithms. Approaches such as compressive chunking which reduces data dimensionality before segmentation and incremental segmentation which processes data in small temporal windows enable real-time operation. Future research should explore edge-compatible methods that leverage hardware acceleration (e.g., GPUs, TPUs) to support efficient deployment in resource-limited environments such as IoT devices and mobile platforms.

**4. Robustness to Noise**

Real-world multimodal data often contains significant noise, making noise-tolerant chunking crucial. Preprocessing techniques like SpecAugment for audio or image enhancement for low-quality visuals can help improve input quality and enhance the robustness of downstream chunking performance. Additionally, algorithms that adapt to noise e.g., probabilistic boundary detection or ensemble-based approaches—can maintain performance under adverse conditions. Future research should integrate robustness into chunking pipelines to ensure reliable operation across heterogeneous data sources.

**5. Standardized Evaluation Benchmarks**

The lack of unified datasets and metrics impedes comparative evaluation of chunking methods. Establishing benchmarks with diverse multimodal content (e.g., documents with text, images, and audio) and metrics assessing alignment, coherence, and task-specific performance (e.g., retrieval accuracy, generation quality) is critical. Existing NLP initiatives, such as GLUE, can serve as inspiration for multimodal chunking benchmarks. Open-source datasets and standardized evaluation protocols will accelerate reproducibility and research progress.

**6. Dynamic and Adaptive Chunking**

Context-aware or query-driven chunking that adapts to task requirements or data characteristics can improve flexibility. Approaches like Mixture-of-Granularity (MoG) select chunk sizes based on query scope, balancing precision and recall [25]. Incorporating real-time feedback, such as user interactions or task performance signals, could further enhance relevance [26]. Exploring dynamic chunking algorithms based on reinforcement learning or online learning could enable systems to continuously adapt to changing data patterns or user requirements, improving flexibility and long-term performance.

**7. Ethical Chunking Strategies**

Ensuring fairness in chunking involves addressing biases present in pretrained models (e.g., CLIP) and training datasets, such as those used for speaker diarization [32]. Techniques like adversarial debiasing, bias auditing, and fairness-aware training can help reduce inequities in chunking outcomes. Future research should also prioritize transparency and explainability in boundary decisions, particularly in sensitive areas like healthcare and legal applications.

**8. Interdisciplinary Applications**

Multimodal chunking can also be applied to emerging domains such as augmented reality (AR), virtual reality (VR), and education. For example, segmenting lecture videos along with slides and transcripts can create more interactive and adaptive learning experiences. Similarly, AR/VR platforms can leverage real-time chunking to enhance immersive analytics. Interdisciplinary research that bridges AI with cognitive science and human-computer interaction can reveal new requirements and foster user-centered chunking solutions.

## 6.3 Conclusion

Multimodal chunking is a core technique that allows AI systems to understand and reason across diverse data types text, images, audio, and video. This survey has provided a comprehensive taxonomy and in-depth analysis of chunking strategies, spanning both unimodal and cross-modal approaches. We examined their theoretical underpinnings, implementation methods, and trade-offs. By organizing these techniques along dimensions such as modality, granularity, heuristics, alignment, and purpose, we offer a clear framework for researchers and practitioners to design, implement, and evaluate effective chunking pipelines.

Despite significant progress, multimodal chunking still faces challenges related to semantic coherence, scalability, robustness, generalization, evaluation, dynamic temporal and spatial variation, and ethical concerns. These challenges highlight how chunking quality directly influences the performance of even the most advanced foundation models. The future directions discussed in this survey including hybrid architectures, self-supervised learning, real-time and noise-tolerant methods, standardized benchmarks, adaptive algorithms, fairness-aware designs, and interdisciplinary applications chart a path forward for addressing these limitations.

Looking ahead, multimodal chunking will remain central to realizing the full potential of AI systems, driving progress in areas such as multimedia retrieval, cross-modal reasoning, and interactive analytics. By tackling current challenges and advancing the proposed research directions, the community can build chunking strategies that are more robust, scalable, and equitable. As multimodal data continues to grow in complexity and diversity, chunking will persist as a foundational element shaping the efficiency, reliability, and overall capability of AI in real-world applications.

# Appendix A: Notation and Symbols

This appendix presents a comprehensive list of mathematical symbols and notations used throughout the paper *"Chunking Strategies for Multimodal AI Systems."* The table is organized alphabetically by symbol and grouped according to application context general, text chunking, image chunking, audio chunking, video chunking, and cross-modal chunking to enhance clarity and ease of reference. Each entry includes the symbol, its category, a brief description of its meaning, and a reference to the section where it is first introduced or primarily applied.

| **Symbol** | **Category** | **Description** | **Section** |
|---|---|---|---|
| $A$ | Audio | Audio signal with total duration or samples N. | 4.3.1 |
| $B_i$ | Image | Bounding box for the i-th object, defined by top-left corner $(x_i, y_i)$, width $w_i$, and height $h_i$. | 4.2.2 |
| $C$ | General | A chunk, representing a segmented unit of data (text, image, audio, video, or multimodal). | 2, 3 |
| $D$ | General | A document or data input, which may contain text, images, audio, or video. | 4.1.3, 7 |
| $e_i$ | Cross-Modal | Multimodal embedding vector combining features from text, image, audio, or video segments. | 7.3 |
| $e_i$ | Text | Embedding vector for the i-th sentence or text segment. | 4.1.2.4 |

| $e_t$ | Audio | Embedding vector for an audio segment at time t. | 4.3.4 |
|---|---|---|---|
| $e_t$ | Video | Multimodal embedding vector for a video clip at time t, combining visual, audio, or transcript features. | 4.4.4 |
| $f$ | General | A function (e.g., alignment, filtering, or similarity function) used in chunking processes. | 4.2.3, 7.1 |
| $G$ | Text | Granularity level in Mixture-of-Granularity (MoG) chunking (e.g., sentence, paragraph). | 4.1.3.4 |
| $g$ | Text | Specific granularity selected by the router in MoG chunking. | 4.1.3.4 |
| $I$ | Image | Input image with height H, width W, and C color channels. | 4.2.1 |
| $K$ | Text | Fixed chunk size (e.g., number of tokens, characters, or words). | 4.1.1 |
| $L$ | General | Number of chunks generated from a given input. | 4.2.2, 4.3.2, 4.4.2, 7.2 |
| $l_i$ | Audio | Label for the i-th chunk (e.g., speaker ID or content type). | 4.3.3, 4.3.4 |
| $l_i$ | Video | Label for the i-th chunk (e.g., shot, action, or semantic theme). | 4.4.2, 4.4.3, 4.4.4 |
| $N$ | General | Total number of tokens, pixels, samples, or frames in the input data. | 4.1.1, 4.2.1, 4.3.1, 4.4.1 |
| $N_i$ | Image | The i-th node in a scene graph, including bounding box and label. | 4.2.5 |
| $O$ | Text | Overlap size in overlapping fixed-size chunking. | 4.1.1.2 |
| $P_i$ | Image | The i-th patch in patch-based chunking, defined by coordinates and size. | 4.2.1 |
| $Q$ | Text | Query input for query-driven chunking. | 4.1.3.1 |
| $R_i$ | Image | The i-th region proposal, defined by bounding box coordinates. | 4.2.3 |
| $S$ | Text | Stride size in sliding window chunking, determining the step between chunks. | 4.1.1.3 |
| $S_i$ | Text | The i-th sentence in a text sequence. | 4.1.2.4 |
| $T$ | Audio | Fixed chunk duration in seconds or samples for fixed-length chunking. | 4.3.1 |
| $T$ | Text | Input text sequence to be segmented. | 4.1.3 |
| $T$ | Video | Fixed chunk duration in seconds or frames for fixed-length chunking. | 4.4.1 |
| $T_t$ | Video | Transcript segment associated with a video clip at time t. | 4.4.4 |
| $t_i^s, t_i^e$ | Audio | Start and end sample indices (or times) for the i-th chunk in silence-based or speaker-based chunking. | 4.3.2, 4.3.3 |

| $t_i^s, t_i^e$ | Video | Start and end frame indices (or times) for the i-th chunk in shot-based or action-based chunking. | 4.4.2, 4.4.3 |
|---|---|---|---|
| τ | General | Threshold parameter for similarity, confidence, or boundary detection. | 4.1.2.4, 4.3.4, 4.4.3 |
| $V$ | Video | Video sequence with total frames F. | 4.4.1 |
| $VAD(x_t)$ | Audio | Voice Activity Detection function, labeling sample $x_t$ as silence (0) or non-silence (1). | 4.3.2 |
| $w_i$ | Text | Word frequency vector for the i-th pseudo-sentence in TextTiling. | 4.1.2.5 |

Caption**:** Notation and symbols used in the paper, sorted alphabetically by symbol and categorized by their application context.

Notes:

- Indexing: Subscripts (e.g., $C_i, P_i$) denote the i-th element in a sequence (e.g., chunk, patch).
- Superscripts: Superscripts (e.g., $t_i^s, t_i^e$) indicate start (s) or end (e) indices or timestamps.
- Embedding Functions: Embeddings (e.g., $e_i, e_t$) are dense vector representations generated by models such as BERT, CLIP, or VideoCLIP, typically used for similarity-based chunking.
- Thresholds: The parameter $\tau$ is context-dependent, representing thresholds for similarity, confidence, or boundary detection. Its optimal value must be tuned according to the task and dataset.